\documentclass[10pt,twocolumn,letterpaper]{article}

\usepackage{cvpr}              

\usepackage{graphicx}
\usepackage{amsmath,bm}
\usepackage{amssymb}
\usepackage{booktabs}
\usepackage{amsfonts}
\usepackage{url}

\usepackage[ruled,linesnumbered]{algorithm2e}
\usepackage{caption}
\usepackage{subcaption}
\usepackage{bbm}
\usepackage{latexsym}
\usepackage{soul}
\usepackage{tabulary,multirow,overpic}
\usepackage{booktabs}
\usepackage{balance}
\usepackage{pifont}
\usepackage{CJKutf8}
\usepackage{comment}
\usepackage[T1]{fontenc}
\usepackage[utf8]{inputenc}

\usepackage[table]{xcolor}
\usepackage[british, english, american]{babel}

\usepackage{listings}
\usepackage[frozencache,cachedir=.]{minted}
\usepackage{color, colortbl}
\newcommand{\x}{{$\times$}}

\DeclareMathOperator*{\argtopk}{arg\,topk}

\newcommand{\xmark}{\ding{55}}%

\newlength\savewidth\newcommand\shline{\noalign{\global\savewidth\arrayrulewidth
		\global\arrayrulewidth 1pt}\hline\noalign{\global\arrayrulewidth\savewidth}}
\newcommand{\tablestyle}[2]{\setlength{\tabcolsep}{#1}\renewcommand{\arraystretch}{#2}\centering\footnotesize}
\renewcommand{\paragraph}[1]{\vspace{1.25mm}\noindent\textbf{#1}}

\newcommand{\app}{\raise.17ex\hbox{$\scriptstyle\sim$}}

\definecolor{defaultcolor}{gray}{0.9}
\usepackage[pagebackref=false, breaklinks=true, letterpaper=true, colorlinks, urlcolor=gray,
citecolor=citecolor, linkcolor=linkcolor, bookmarks=false]{hyperref}
\definecolor{citecolor}{HTML}{0071BC}
\definecolor{linkcolor}{HTML}{ED1C24}

\usepackage[capitalize]{cleveref}
\crefname{section}{Sec.}{Secs.}
\Crefname{section}{Section}{Sections}
\Crefname{table}{Table}{Tables}
\crefname{table}{Tab.}{Tabs.}

\def\cvprPaperID{x} 
\def\confName{CVPR}
\def\confYear{2022}

\title{CiT: Curation in Training for Effective Vision-Language Data}

\author{Hu Xu\qquad Saining Xie\qquad Po-Yao Huang\qquad Licheng Yu\quad\\ Russell Howes\quad Gargi Ghosh\quad Luke Zettlemoyer\quad Christoph Feichtenhofer\\
FAIR, Meta AI\\
\href{https://github.com/facebookresearch/CiT}{\color{purple}{https://github.com/facebookresearch/CiT}}\\
}

\begin{document}

\maketitle

\begin{abstract}
\vspace{-5pt}
Large vision-language models are generally applicable to many downstream tasks, but come at an exorbitant training cost that only large institutions can afford.
This paper trades generality for efficiency and presents Curation in Training (CiT), a simple and efficient vision-text learning algorithm that couples a data objective
into training.
CiT automatically yields quality data to speed-up contrastive image-text training and alleviates the need for an offline data filtering pipeline, allowing broad data sources (including raw image-text pairs from the web). 
CiT contains two loops: an outer loop curating the training data and an inner loop consuming the curated training data.
The text encoder connects the two loops.  
Given metadata for tasks of interest, 
e.g., class names, and a large pool of image-text pairs, CiT alternatively selects
relevant training data from the pool by measuring the similarity of their text embeddings and embeddings of the metadata. 
In our experiments, we observe that CiT can speed up training by over an order of magnitude, especially if the raw data size is large. 
\vspace{-5pt}

\end{abstract}

\section{Introduction}
\label{sec:intro}
Vision-language models have demonstrated success for fine-tuning and zero-shot transfer to downstream tasks\cite{radford2021learning,jia2021scaling,singh2021flava} by training on a general-purpose large-scale dataset instead of a small task-level dataset.
While general, large-scale pre-training is computationally expensive (\eg CoCa\cite{yu2022coca} trains on 2048 TPUs for 5 days) and typically performed on a \textit{pre-filtered} dataset (\eg WIT400M~\cite{radford2021learning} used by CLIP~\cite{radford2021learning} is created by searching for image-text pairs with text containing a set of 500,000 queries and \cite{schuhmann2021laion} uses this model to create another dataset).

\begin{figure}[t!]
\centering
\vspace{-5pt}
\includegraphics[width=1.0\linewidth]{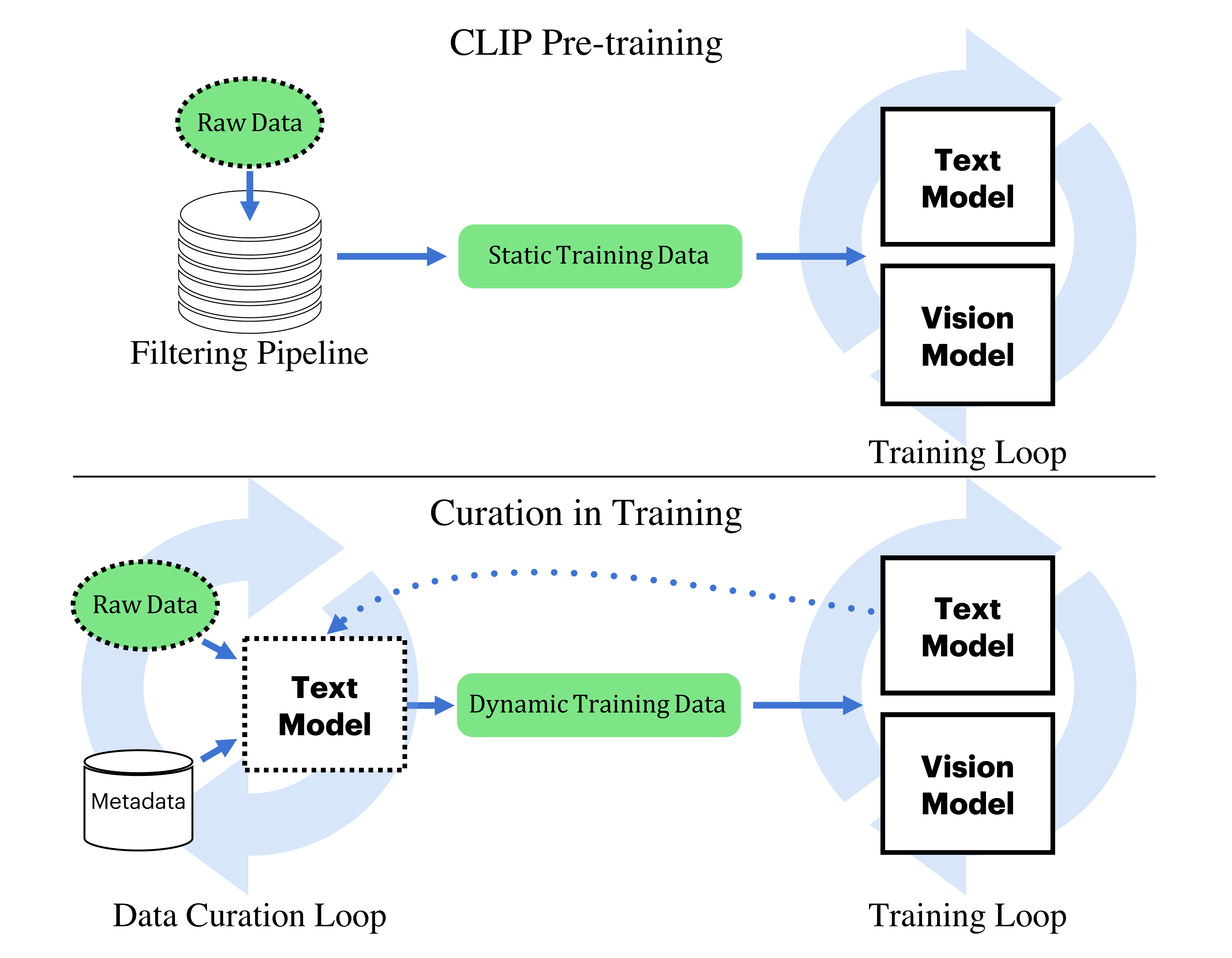}
\vspace{-20pt}
\caption{A conceptual illustration of CLIP training \vs CiT.
Vanilla CLIP training uses static data from offline human filtering (\eg cleaned YFCC15M or WIT400M~\cite{radford2021learning}) and optimizes the model.
Instead, our CiT incorporates dynamic data curation into training in two loops: (\textit{i}) an outer curation loop improving data (for downstream tasks) given the current model; (\textit{ii}) an inner loop optimizing the model given the curated data.
The trained text model connects the loops by providing embeddings for curation.
}

\label{fig:concept}
\vspace{-15pt}
\end{figure}

Such filtering pipelines usually involve manual labor-intensive efforts to remove data that is unlikely useful for downstream tasks~\cite{jia2021scaling,radford2021learning}.  
Recent effort has been made to curate data for high-quality image-text pairs (such as CC3M\cite{sharma-etal-2018-conceptual}, CC12M\cite{changpinyo2021conceptual}, YFCC15M\cite{thomee2016yfcc100m,radford2021learning}, WIT400M\cite{radford2021learning} and LAION\cite{schuhmann2021laion,schuhmannlaion}). Nevertheless, 
research is typically \textit{tied} to the static datasets or model weights (if the data is not released) and is not able to access or change the data pipelines or model architectures. Further, work is \textit{limited} by the prohibitive cost of training on these large image-text datasets (\eg the CLIP model is trained on WIT400M for 12 days using 256 GPUs).

In this work, our goal is to empower training with the
capability of adjusting the data distribution. 
Our intention is to dynamically curate the data during training and our key idea is to use the learned text representation of vision-language models to measure relevance of the data \wrt the task of interest. Given metadata (from downstream tasks \eg a class name such as ``chicken''), we measure its embedding similarity to the training data. This similarity can guide us for the decision of including this data into our training process. For example a caption containing the word ``giraffe'' will have higher embedding similarity to ``chicken'' than a caption such as ``throwback Thursday''.

Driven by this idea, we presents a simple algorithm that incorporates data Curation in Training (CiT), 
aiming at improving both data efficiency and model performance.
CiT works as follows.
Given a large source of image-text pairs and metadata
(\eg a list of class names used in this paper), CiT alternatively performs curation of the data and training on that curated data.
As shown in Figure \ref{fig:concept}, 
CiT contains two loops: an outer loop to curate data given the current model and an inner loop trains the model given the curated data. 
Similar as Locked image Tuning (LiT~\cite{zhai2022lit}), CiT uses pre-trained image and text encoders and freezes the image one. 
The text model connects the two loops by serving curated data to inner loop for training which in turn learns good representations for the outer loop for curation. 

CiT can speed up training by multiple orders of magnitude, especially if the raw data size is large; \eg when trained on LAION-400M data, CiT reaches similar ImageNet zero-shot\footnote{Zero-shot refers to not seeing any training examples of the target dataset. We note that our approach uses extra information of the downstream task, such as class names; however, this metadata is easy to acquire and can be of various forms as shown in experiments. } accuracy as OpenCLIP \cite{wortsman2022robust},  while being 37.7$\times$ faster in training.  
Since CiT changes the training data distribution that focuses on one or more tasks of interest, it can even handle image-text pairs from any (noisy) source with unknown distribution.
Our experiments reveal that vanilla CLIP/LiT training fails on \textit{raw} random image-text pairs crawled from the web, while CiT trains easily.

\section{Related Work}
\label{sec:related}
\paragraph{Vision-Language Learning.}
Contrastive learning was initially popular in vision self-supervision\cite{wu2018unsupervised,chen2020simple,he2020momentum} and later adopted for cross-modal learning\cite{radford2021learning,miech2019howto100m,miech2020end,xu2021videoclip,li2021supervision}.
CLIP\cite{radford2021learning} populates the idea of contrastive learning from image-text pairs (used before \eg in ConVIRT\cite{zhang2020contrastive}) at scale and shows a strong performance of zero-shot transfer to image classification and retrieval tasks.
SLIP\cite{mu2021slip} combines image self-supervision and language supervision. 
LiT\cite{zhai2022lit} shows that when a good pre-trained vision encoder is adopted, it is better to lock (freeze) the well pre-trained vision encoder to protect vision representations from being corrupted by noisy language supervision.
Flamingo also use pre-trained models for various tasks\cite{alayrac2022flamingo}. 

\paragraph{Vision-Language Data.}
Large-scale vision-language learning is typically coupled to a data pipeline to yield high-quality data for efficient training\cite{singh2021flava,yuan2021florence,jia2021scaling}.
For example, CC3M\cite{sharma-etal-2018-conceptual} heavily filters web crawled pairs and only keeps 0.1\% of the raw data. Both CC3M and CC12M\cite{changpinyo2021conceptual} leverage Google Cloud APIs with models predicting a large number of classes (on the order of $10^5$)\cite{sharma-etal-2018-conceptual} to filter out mismatched image-text pairs.
YFCC100M is curated from Yahoo Flicker using text fields (such as title, description, etc.). This ensures certain data quality but limits the scale. Later YFCC100M is further cleaned as YFCC15M to contain English-only image-text pairs by \cite{radford2021learning}.
Due to the limited scale, CLIP further curates a WebImageText dataset (WIT400M) by formulating queries from Wikipedia and performing searching them online to obtain image-text pairs.
Florence\cite{yuan2021florence} curates a dataset with the extra multi-label signals to improve supervision.
ALIGN\cite{jia2021scaling} relaxes CC12M filtering to show that training on 1.8B noisy pairs can achieve CLIP-level performance.
FLAVA\cite{singh2021flava} combines existing human annotated datasets of smaller scale for high-quality image-text pairs.
Different to related research, CiT improves data \textit{within} the training algorithm, and not as a pre-filtering. We demonstrate that such approach allows us to effectively learn from \textit{raw} image-text pairs.

\paragraph{Related Areas.}
Our work is related to research in other domains. In NLP, there are existing works on domain-adaptive finetuning and retrieval \cite{xu2019bert,zhang2019curriculum,gururangan2020don,li2020task,lee2020biobert,yates2021pretrained}. In machine learning research, subset selection~\cite{wei2015submodularity,killamsetty2021glister} cast data selection as a discrete bi-level optimization problem. 

\section{Method}
\label{sec:method}

In CLIP pre-training, the training objective (\eg contrastive image-text correspondence) operates as a training proxy that approximates downstream tasks (\eg  classification accuracy). Our CiT introduces a \textit{data proxy} to fit the \textit{data distribution} to downstream tasks. 
In this section, we first go through the details of the CiT algorithm in \S\ref{sec:algo}, training loop in \S\ref{sec:training} and the data proxy for the curation loop in \S\ref{sec:curation}.

\subsection{CiT Algorithm}
\label{sec:algo}
CiT contains two loops:
the curation loop curates data given the current weights of the model and the training loop optimizing the weights given the curated data.

Let $\mathcal{D}=\{(x_\text{img}^i, x_\text{txt}^i)\}^N_{i=1}$, be the set of source of image-text pairs.
Then $\mathcal{D}_C \subseteq \mathcal{D}$ is the actual \textit{training data} we aim to curate from the source.
We define two functions:
(i) $\textit{Curation}(\mathcal{D}; \Theta)$, and (ii) $\textit{Training}(\Theta; \mathcal{D}_T)$, for curation and training loops, respectively.
Importantly, the weights of the learned model $\Theta$ connects the two loops and serves the curation loop with the updated representations from the training loop.
There is no notion of a \textit{fixed dataset} or training epochs over $\mathcal{D}$; instead, we view the data source as an online \textit{data stream}.
CiT uses a sequential setup that alternatively performs curation for every $s$ pairs of training.

CiT is shown in Algorithm \ref{alg:cit}.
It takes 3 inputs: a data source $\mathcal{D}$, the pre-trained weights $\Theta$ and a training budget $b$, which can be training time, resources consumed, \etc We simply use steps of weight updates as the training cost in this paper. Line 1 initializes the training budget.
Line 2 determines if current training exceeds that training budget.
The main framework of CiT is to alternatively perform curation and training in line 2-4. 
To recap CLIP pre-training, we first detail the training function next.

\subsection{Training}
 \label{sec:training} 
The core of CLIP~\cite{radford2021learning} training is the contrastive cross-modal objective serving as the proxy to approximates downstream tasks (\eg higher classification accuracy).
This objective pulls embeddings of positive image-text pairs closer and pushes negative pairs from other examples in a training batch apart; thus it creates a proxy for classification, which has one example per class and the rest of the batch are other classes described by natural language.

The training loop is shown in Algorithm \ref{alg:train}, with the training data $\mathcal{D}_C$, delivered from curation. 
We let $m(\cdot; \cdot)$ denote the image-text model. 
We use $\mathrm{sim}(\mathbf{x}_\text{img}, \mathbf{x}_\text{txt}) = \mathbf{x}_\text{img} \mathbf{x}^\top_\text{txt} / (\lVert \mathbf{x}_\text{img}\rVert \lVert \mathbf{x}_\text{txt}\rVert)$ in line 3 to compute the image-to-text cosine similarity, divided by a trainable temperature $\tau$.
Our CiT training objective has almost the same structure as in CLIP, except that we only use an image-to-text (and no text-to-image) contrastive loss ($\mathcal{L}_\text{img2txt}$) in line 4.
We ablate this loss versus the averaged bidirectional contrastive loss (used by CLIP) in our experiments.
Line 5 updates the model parameters and line 6 counts training cost. 

\begin{algorithm}[!tp]
\fontsize{9pt}{9pt}\selectfont
\DontPrintSemicolon
\caption{CiT (see \S\ref{sec:impl:pseudo} for pseudo code)}
\label{alg:cit}
\SetKwInOut{Input}{Input}
\Input{$\mathcal{D}$: data source\\$\Theta$: model's  pre-trained weights\\$b$: training budget}
$c \gets 0$ \;
\While{$c < b$}{
    $\mathcal{D}_T \gets \textit{Curation}\big(\mathcal{D};  \Theta\big)$ \;
    $\Theta, n \gets \textit{Training}(\Theta; \mathcal{D}_T)$ \;
    $c \gets c + n$ \;
}

\end{algorithm}

\begin{algorithm}[!t]
\fontsize{9pt}{9pt}\selectfont
\DontPrintSemicolon
\caption{Curation}
\label{alg:curation}
\SetKwInOut{Input}{Input}
\SetKwInOut{Constant}{Constant}
\Input{$\Theta$: model's current weights\\$\mathcal{D}$: data source}
\Constant{$m(\cdot; \cdot)$: model architecture\\$\mathcal{T}_\text{meta}$: metadata for tasks of interests\\$s$: number of expected pairs}

$\mathbf{x}_\text{meta} \gets m(\mathcal{T}_\text{meta}; \Theta)$ \;
$\mathcal{D}_C \gets \varnothing$ \;
\While{|$\mathcal{D}_C$| < $s$\text{ and }$\mathcal{D}_\text{raw} \subset \mathcal{D}$}{
    $\mathcal{D}_\text{raw,txt} \gets \{x_\text{txt}^i|(x_\text{img}^i, x_\text{txt}^i) \in \mathcal{D}_\text{raw}$\} \;
    $\mathbf{x}_\text{txt} \gets m(\mathcal{D}_\text{raw,txt}; \Theta)$ \;
    $f \gets \textit{DataProxy}(\mathcal{D}_\text{raw}, \mathbf{x}_\text{txt}, \mathbf{x}_\text{meta})$ \;
    $\mathcal{D}_C \gets f(\mathcal{D}_\text{raw};\Theta, \mathcal{D}_\text{raw}, \mathcal{T}_\text{meta}) \cup \mathcal{D}_C$ \;
}
\Return $\mathcal{D}_C$ \;
\end{algorithm}

\begin{algorithm}[!t]
\fontsize{9pt}{9pt}\selectfont
\DontPrintSemicolon
\caption{Training}
\label{alg:train}

\SetKwInOut{Input}{Input}
\SetKwInOut{Constant}{Constant}
\Input{$\mathcal{D}_C$: curated training data\\$\Theta$: model's weights}
\Constant{$m(\cdot; \cdot)$: model architecture}

\ForEach{$\mathcal{D}_\text{batch} \subset \mathcal{D}_C$ }{
    $\mathbf{x}_\text{img}, \mathbf{x}_\text{txt} \gets m(\mathcal{D}_\text{batch}; \Theta)$ \;
    $l \gets \mathrm{sim}(\mathbf{x}_\text{img}, \mathbf{x}_\text{txt})/\tau$ \;
    $\mathcal{L}_\text{img2txt} \gets \textit{CrossEntropy}(l, \textit{arange}(|\mathcal{D}_\text{batch}|))$ \;
    $\Theta \gets \mathcal{L}_\text{img2txt}(\Theta; \mathcal{D}_\text{batch})$ \;
    $n \gets n + 1$ \;
}
\Return $\Theta$, $n$ \;
\end{algorithm}

\subsection{Curation}
\label{sec:curation}

CiT also has a data objective that curates data using the (previously updated) model. Encoding the data with an updated model allows for better representation of the data. 
Akin to the contrastive objective for training, the core function in curation is a \textit{data proxy} (or objective) that selects data based on the metadata (\eg a list of class names).

We detail the curation loop in Algorithm \ref{alg:curation}.
It takes the following inputs: model weights $\Theta$, a data source $\mathcal{D}$, the model architecture, the metadata for downstream tasks $\mathcal{T}_\text{meta}$ and an expected size of curated data $s$.
$\mathcal{T}_\text{meta}$ is a list containing a pre-defined taxonomy; (\eg ImageNet WordNet lemmas or a combination from a group of tasks in our experiments), but could be generalized to other forms of text.

Algorithm \ref{alg:curation} first obtains the embeddings for the metadata in line 1.
Then it sets up the curated set $\mathcal{D}_C$ for the next round of training and keeps curating data in line 3-7.
Line 3 gets the next batch of raw image-text pairs. Line 4 obtains its text part and line 5 computes the text embedding from the current model.
Line 6 is the \textit{data proxy}, which approximates the data distribution for the downstream tasks (detailed in the next subsection).
Lastly, we merge the newly curated subset into the curated set $\mathcal{D}_C$.

\paragraph{Data Proxy.}
\label{sec:curation_proxy}
We use language-based metadata and the text encoder to measure the relevance of training data. This favors efficiency because the text encoders are typically \textit{significantly cheaper} to evaluate (\eg the text encoder only uses \app4.6\% of the \mbox{ViT-L} image-encoders' compute).

In $\textit{DataProxy}(\mathcal{D}_\text{raw}, \mathbf{x}_\text{txt}, \mathbf{x}_\text{meta})$ of Algorithm \ref{alg:curation}, we first compute the  similarities of text embeddings ($\mathbf{x}_\text{txt}$) over embeddings of the metadata ($\mathbf{x}_\text{meta}$):
\begin{equation}
\begin{split}
\label{eq:sim}
v^i_\text{max} = \max_{j}(\mathrm{sim}(\mathbf{x}^i_\text{txt}, \mathbf{x}^j_\text{meta})),
\end{split}
\end{equation}
where 
$\mathrm{sim}(\mathbf{x}^i_\text{txt}, \mathbf{x}^j_\text{meta}) = \mathbf{x}^i_\text{txt} \mathbf{x}^{j,\top}_\text{meta} / (\lVert \mathbf{x}^i_\text{txt}\rVert \lVert \mathbf{x}^j_\text{meta}\rVert)$ is the cosine similarity between embeddings of sample $i$ and metadata $j$.
Here the highest similarity over all metadata $v_\text{max}^i$ is used to measure the sample quality. 

Let $\mathcal{D}_t=\{(x_\text{img}^i, x_\text{txt}^i)|(x_\text{img}^i, x_\text{txt}^i) \in \mathcal{D}_\text{raw} \text{ and } v_\text{max}^i > t\}$ denote a subset, where all samples have a maximum similarity above a curation threshold $t$. Given the best possible match to metadata, we use a mixed strategy to determine if a sample shall be used:
\begin{equation}
\begin{split}
\label{eq:policy}
\begin{cases}
\mathcal{D}_t &\text{if}\ \frac{|\mathcal{D}_t|}{|\mathcal{D}_\text{raw}|} > \gamma, \\
\argtopk_{i}(v_\text{max}^i, k=\gamma |\mathcal{D}_\text{raw}|), &\text{otherwise,}
\end{cases}
\end{split}
\end{equation}
where $\frac{|\mathcal{D}_t|}{|\mathcal{D}_\text{raw}|}$ is the \textit{ratio of curation} with $\gamma$ being a pre-defined \textit{minimal} ratio of curation.
If enough samples meet the threshold $t$, $\mathcal{D}_t$ is used.
Otherwise, we use a \textit{minimal ratio} $\gamma$ of samples, that represent the top-$k$ matching ones (with $k=\gamma |\mathcal{D}_\text{raw}|$) in terms of similarity across metadata.

The threshold $t$ is crucial for CiT to balance the tradeoff between data quality and quantity.
A higher $t$ leads to high data quality, but can lead a lower ratio of curation.
We adopt this mixed strategy because line 3 in Algorithm \ref{alg:curation} could become a near infinite loop if the ratio of curation is low and not enough data that meets $t$ can be found.
This could happen because the threshold is set too high, or the data source has low metadata correspondence.
The \textit{otherwise} part in equation \ref{eq:policy} resolves this by selecting the $\gamma$ (typically set to around 1\% - 5\%) best possible matches for training. 
See \S\ref{sec:impl:pseudo} for PyTorch pseudo code of CiT.

\section{Experiments}
\label{sec:exp}

We use training data from two categories shown below; clean data that involves human-based offline filter pipelines and raw data that has not undergone cleaning.

\subsection{Cleaned Training Data}
\paragraph{YFCC15M.} We use the 15M subset of YFCC100M\cite{thomee2016yfcc100m} (filtered by \cite{radford2021learning}) as the main evaluation dataset as it is widely adopted in existing literatures\cite{radford2021learning,mu2021slip,zhai2022lit,santurkar2022caption}. It consists of English-only titles, descriptions, and tags. We simply refer to this as YFCC15M in this paper. Except for applying the script from \cite{mu2021slip} to remove HTML formatting, we do not perform any extra filtering or preprocessing. In contrast, LiT\cite{zhai2022lit} performs extra filtering such as removing titles that start with ``DSC'', ``IMG'' and ``Picture'', or removing them if more than half of them contain digits.

\paragraph{CC12M.} Since YFCC15M may lack enough training data, LiT\cite{zhai2022lit} also combines YFCC15M with Conceptual Captions 12M (CC12M) \cite{changpinyo2021conceptual}, which is filtered and transformed from image \& alt-text pairs from web pages. CC12M involves cleaning by supervised models from Google Cloud APIs to match the image's prediction over classes with text.

\paragraph{LAION400M} \cite{schuhmann2021laion} contains 400M English only image-text pairs. It is crawled from \footnote{\url{https://commoncrawl.org}} and later filtered by a CLIP\cite{radford2021learning} model. Thus, LAION400M implicitly carries the data filter pipeline of WIT400M on which CLIP has been trained.
\subsection{Raw Training Data}

\paragraph{YFCC100M}. We use the raw YFCC100M (the source of YFCC15M) to compare with YFCC15M. Note that YFCC100M is multilingual, whereas YFCC15M is English.

\paragraph{Raw Image-Text Crawl.} To challenge CiT with real-world data, we further collect raw (unfiltered) image-text pairs from Common Crawl.
We only perform de-duplication and NSFW filtering, but \textit{no} filtering on image-text association. 
This ended with 1.2B multilingual image-text pairs and 28.56\% pairs are English (identified by our language identification system but this information is not used for CiT training). 
As such, about 343M image-text pairs are in English, which are slightly smaller than the scale of WIT400M or LAION400M, but much more noisy.

\subsection{Implementation and Training} 
Our training recipe  uses
a global batch size of 16,384, which is trained in 16 Nvidia V100 32GB GPUs.
Our vision encoder corresponds to ViT \cite{dosovitskiy2020image} of various sizes and the text encoder defaults to $\text{BERT}_\text{base}$-SimCSE \cite{devlin-etal-2019-bert,gao2021simcse} with a maximum token length of 32, similar to LiT~\cite{zhai2022lit}.
Unless specified, we set a budget of training to be within $b=5000$ steps (81M image-text pairs).
We report hyper-parameters and an extra low-cost single-GPU setting in  \S\ref{sec:impl}.

We use pre-trained vision and text encoders and join them via two randomly initialized projection layers. Following LiT, we freeze the vision encoder and make the text encoder and two projection layers trainable. 
One can either use the text representation \textit{before}, or \textit{after} the projection layer for computing cosine similarity during curation. We ablate these two choices in \S\ref{sec:ablations}.

\begin{table*}[t]\centering
\vspace{-10pt}
\tablestyle{5pt}{1.1}
\subfloat[{Curation effect}]{
	\begin{tabular}{l l}
		Curation
		& Acc\\
		\toprule
		\rowcolor{defaultcolor} online & \textbf{61.4} \\
		offline  & 57.5 \\
		no  & 53.8 \\
		&\\
    	&\\
	\end{tabular}
	\label{tbl:ablation:stage}
}
\hspace{3mm}
\hspace{3mm}
\subfloat[{Curation freq.}]{
	\begin{tabular}{l r}
		\# of steps
		& Acc\\
		\toprule
		 50 & 61.0 \\
		\rowcolor{defaultcolor} 100 & 61.4 \\
		200 & \textbf{61.5} \\
		300 & 61.1 \\
		&\\

	\end{tabular}
	\label{tbl:ablation:freq}
}
\hspace{3mm}
\subfloat[{Curation feature}]{
	\begin{tabular}{l r}
		Feature of Curation
		& Acc\\
		\toprule
		\rowcolor{defaultcolor} pooled encoder & \textbf{61.4} \\
		projection output & 60.7 \\
		w/ prompts & 61.4 \\
		&\\
		&\\
	\end{tabular}
	\label{tbl:ablation:filterfeat}
}
\subfloat[{Threshold $t$}]{
	\begin{tabular}{l l}
		Threshold $t$
		& Acc\\
		\toprule
		0.5 & 60.9\\
		\rowcolor{defaultcolor} $t=0.55$ & \textbf{61.4} \\
		0.6 & 61.1\\
		0.7 & 59.7\\
		&\\
	\end{tabular}
	\label{tbl:ablation:threshold}
}
\hspace{3mm}
\subfloat[Text variants]{
	\begin{tabular}{l r}
		Text Variants
		& Acc \\
		\toprule
		\rowcolor{defaultcolor} BERT max len. 32 & \textbf{61.4} \\
		BERT max len. 77 & 61.2 \\
		w/o YFCC tag & 59.1 \\
		w/o YFCC tag aug. & 60.8 \\
		BERT first 6 layers & 60.2 \\
	\end{tabular}
	\label{tbl:ablation:text}
}
\vspace{-5pt}
\caption{\textbf{Ablation experiments}. We use MoCo-v3 / $\text{BERT}_\text{base}$-SimCSE, YFCC15M as data source and report IN-1K Accuracy.}
\label{tbl:ablation}
\vspace{-1.em}
\end{table*}

\subsection{Evaluation} 
We evaluate zero-shot (0-shot) transfer \textit{accuracy} of CiT on \textbf{26}  benchmarks, following~\cite{radford2021learning,mu2021slip}.
In our ablation studies, we use YFCC15M as the main data source for training and ImageNet-1K (IN-1K) as the downstream task.
We use prompts from CLIP for all 26 tasks and additionally use the extra 2 prompts from LiT\cite{zhai2022lit} for ImageNet for a fair comparison with LiT.
Following CLIP, we perform prompt ensembling by averaging the class embeddings for each class across the prompt templates. 
For classification, cosine similarity is computed between an image embedding and the averaged class embeddings and the class with the highest cosine similarity is CiT's prediction.
We perform validation every 500 training steps and stop training if the accuracy does not increase over the previous validation. The corresponding total training time (including curation and training) is reported along with the validation accuracy.
We estimate the training time of baselines by re-running them under the same setup as CiT (\ie 16 GPUs) and maximize the GPU usage for best throughput. More results are in \S\ref{sec:extra_results}.

\subsection{Choice of Pre-trained Models} 
We first study the effects of pre-trained encoders.

As \textit{vision encoder}, we consider (1) ViT-B/16~\cite{dosovitskiy2020image} (patch size of 16$\times$16 pixels) with pre-trained weights from self-supervised MoCo-v3 \cite{chen2021empirical},  DINO\cite{caron2021emerging} and  MAE \cite{he2021masked}, all trained on IN-1K but without any labels. To be consistent with LiT\cite{zhai2022lit}, we also consider (2) supervised ViT(AugReg)\cite{steiner2021train} B/32, B/16, and L/16 trained on ImageNet-21K\footnote{We follow LiT here, but note that using IN-21K is not strictly a zero-shot setting, because 999 of the 1000 classes in IN-1K are in IN-21K.}. Finally, we also explore weakly-supervised ViT-B/16 and ViT-H/14 SWAG\cite{singh2022revisiting}.

\begin{table}[!h]
	\vspace{-5pt}
	\centering
	\scalebox{0.72}{\begin{tabular}{l c c c}
			Vision Model
			& Pre-train Obj. 
			& Pre-train Data
			& IN-1K Acc.\\
			\shline
			\rowcolor{defaultcolor} MoCo-v3\cite{chen2021empirical} & Contrastive & IN-1K & 61.4\\
			DINO\cite{caron2021emerging} & Contrastive & IN-1K & 60.3\\
			MAE\cite{he2021masked} & Masking & IN-1K & 42.4 \\
			\hline
			AugReg\cite{steiner2021train} & Supervised & IN-21K & \textbf{69.4} \\
			SWAG\cite{singh2022revisiting} & Weakly-Supervised & IG 3.6B & 67.5 \\
			\hline
	\end{tabular}}
	\vspace{-5pt}
	\caption{Ablation study of different vision encoders on ViT-B/16 with text encoder as $\text{BERT}_\text{base}$-SimCSE on YFCC15M. Pre-training objective matters for CiT training.}
\label{tbl:img:ablation_models}
\vspace{-5pt}
\end{table}

Results for different vision encoder weights under the same ViT-B/16 architecture are in Table \ref{tbl:img:ablation_models}. We notice that the accuracy of MoCo-v3 (61.4\%) and DINO (60.3\%) pre-training are close, while MAE is worse (42.4\%), presumably because the representations learned by instance discrimination (MoCo-v3 and DINO), which learns different embeddings for different images, is closer to zero-shot classification than MAE's training objective. 
AugReg performs best with 69.4\% accuracy, presumably because the supervised pre-training on IN-21K is superior to unsupervised IN-1K pre-training. Finally, SWAG is worse than AugReg, but better than MoCo-v3. In the following experiments of this section, we will show larger variants.

For \textit{text encoder}, we consider self-supervised base models from (1) language models BERT \cite{devlin-etal-2019-bert};
and contrastive tuned (2) BERT-SimCSE and RoBERTa-SimCSE \cite{gao2021simcse}, as shown in Table \ref{tbl:txt:ablation_models}.

\begin{table}[h!]
\vspace{-5pt}
\centering
\scalebox{0.78}{\begin{tabular}{l l c}
		Text Model & Pre-training obj. & IN-1K Acc.\\
		\shline
		$\text{BERT}_\text{base}$(uncased)\cite{devlin-etal-2019-bert} & from scratch & 57.7 \\
		\rowcolor{defaultcolor} $\text{BERT}_\text{base}$(uncased)\cite{devlin-etal-2019-bert} &  SimCSE\cite{gao2021simcse} & \textbf{61.4}\\
		$\text{BERT}_\text{base}$(uncased)\cite{devlin-etal-2019-bert} & BERT NSP\cite{devlin-etal-2019-bert} & 59.9 \\
		$\text{RoBERTa}_\text{base}$\cite{liu2019roberta} & SimCSE\cite{gao2021simcse} & 59.7\\
		\hline
\end{tabular}}
\vspace{-5pt}
\caption{Ablation study of different text encoders with MoCo-v3 on YFCC15M: contrastive pre-training yields better accuracy.}
\label{tbl:txt:ablation_models}
\vspace{-5pt}
\end{table}

We observe similar trends as for vision: SimCSE trained 
BERT is better than vanilla BERT or RoBERTa, probably because contrastively trained \texttt{[CLS]} token by SimCSE can perform better text similarity than BERT's pairwise (a.k.a, next sentence prediction) trained \texttt{[CLS]} token or RoBERTa's no training on \texttt{[CLS]} token.

\subsection{Ablations} \label{sec:ablations}
We adopt the combination of MoCo-v3 ViT B/16 and BERT-SimCSE as our default setting.
We summarize ablation experiments of CiT in Table \ref{tbl:ablation}.

\paragraph{Stage of Curation.} We first ablate the effects of curation in Table \ref{tbl:ablation:stage}. We see that CiT has a \textbf{7.6}\% boost compared to \textit{no curation}. We further ablate a \textit{offline} curation before training. This is sub-optimal as the SimCSE purely pre-trained from the text may not learn good representations for semantic-level similarity
(discussion in \S\ref{sec:algo}). 

\paragraph{Frequency of Curation.} Next, we are interested in how frequently curation needs to be performed. Table \ref{tbl:ablation:freq} varies the number of steps (and therefore pairs $s$ when multiplied with the batch-size) for curation (in Alg.~\ref{alg:curation}). 
We found that curating too frequent or infrequent yields sub-optimal results, but the change is marginal so we chose 100 steps as default. 

\paragraph{Feature for Curation.} 
In Table \ref{tbl:ablation:filterfeat}, we find that using the feature before the projection layer (\eg the direct output of SimCSE) is better than the features from the projection layer. This is probably because the projection layer tends to be more unstable during training (\eg randomly initialized and needs longer training to align with the visual representation), whereas the SimCSE embedding is already pre-trained for text similarity.

\paragraph{Threshold.}  In Table \ref{tbl:ablation:threshold} we ablate the threshold $t$, which controls the trade-off for data quality and quantity. A lower threshold adds more low-quality data and a higher threshold reduces data quantity, so $t=0.55$ is a good balance.

\paragraph{Text Variants.} We ablate the length of text encoders in Table \ref{tbl:ablation:text} to understand the memory/text sequence length tradeoff. We find that longer text sequences (77) (we reduce batch size per GPU to half and double the number of GPUs) are slightly worse.
We also ablate the effectiveness of YFCC15M tag augmentation, adopted from LiT. 
Lastly, we are wondering if a shallow (6 layers) BERT-SimCSE is also a good text encoder. We obtain 1.2\% worse results.

\begin{table*}[!h]
\tablestyle{4pt}{1.0}
\centering
\begin{tabular}{l l l l c l l}
		Pre-train Data & Method & Vision Encoder & Vision Initialization & w/ Labels & Total Time & 
        IN-1K Acc\\
		\shline

	    \multirow{12}{*}{YFCC15M} & CLIP\cite{radford2021learning} & ResNet-50 & scratch & \xmark & 25 hrs & 
        31.3 \\
		& OpenCLIP\cite{wortsman2022robust} & ResNet-50 & scratch & \xmark & 25 hrs &
        32.7 \\
		& LiT\cite{zhai2022lit} & ViT-B/16 & DINO\cite{caron2021emerging} & \xmark & n/a & 
        55.4\\
		& LiT\cite{zhai2022lit} & ViT-B/16 & MoCo-v3\cite{chen2021empirical} & \xmark & n/a & 
        55.5\\
		& LiT\cite{zhai2022lit} & ViT-B/16 & AugReg\cite{steiner2021train} & IN-21K & n/a & 
        55.9\dag\\
		& LiT\cite{zhai2022lit} & ViT-B/32 & AugReg\cite{steiner2021train} & IN-21K & 64 hrs & 
        59.9*\\
		& \textbf{CiT}  & ViT-B/16 & DINO\cite{caron2021emerging} & \xmark & n/a & 
        \textbf{60.3}\\
		& \cellcolor{defaultcolor}\textbf{CiT}  & \cellcolor{defaultcolor}ViT-B/16 & \cellcolor{defaultcolor}MoCo-v3\cite{chen2021empirical} & \cellcolor{defaultcolor}\xmark & \cellcolor{defaultcolor}5 hrs &
        \cellcolor{defaultcolor}\textbf{61.4}\\

		& \textbf{CiT}  & ViT-B/32 & AugReg\cite{steiner2021train} & IN-21K & 11 hrs & 
        \textbf{63.3}\\
		& \textbf{CiT} & ViT-B/16 & AugReg\cite{steiner2021train} & IN-21K & 8 hrs & 
        \textbf{69.4}\\
		& \textbf{CiT}  & ViT-L/16 & AugReg\cite{steiner2021train} & IN-21K & 8 hrs & 
        \textbf{72.0}\\
		& \textbf{CiT}  & ViT-H/14 & SWAG\cite{singh2022revisiting} & IG hashtags & 11 hrs & 
        \textbf{73.7}\\
		\hline
		\multirow{2}{*}{YFCC15M+CC12M} & LiT\cite{zhai2022lit} & ViT-L/16 & AugReg\cite{steiner2021train} & IN-21K & 112 hrs & 
        72.2*\\
		& \textbf{CiT}  & ViT-L/16 & AugReg\cite{steiner2021train} & IN-21K & 32 hrs &  
        \textbf{75.6}\\
		\hline
		\multirow{6}{*}{YFCC100M} & LiT\cite{zhai2022lit} & ViT-B/32 & AugReg\cite{steiner2021train} & IN-21K & 153 hrs & 
        58.9*\\
		& \cellcolor{defaultcolor}\textbf{CiT}  & \cellcolor{defaultcolor}ViT-B/16 & \cellcolor{defaultcolor}MoCo-v3\cite{chen2021empirical} & \cellcolor{defaultcolor}\xmark & \cellcolor{defaultcolor}48 hrs & 
        \cellcolor{defaultcolor}\textbf{64.6}\\
		& \textbf{CiT}  & ViT-B/32 & AugReg\cite{steiner2021train} & IN-21K & 64 hrs & 
        \textbf{65.6}\\
		& \textbf{CiT}  & ViT-B/16 & AugReg\cite{steiner2021train} & IN-21K & 66 hrs & 
        \textbf{72.2}\\
		& \textbf{CiT}  & ViT-L/16 & AugReg\cite{steiner2021train} & IN-21K & 66 hrs & 
        \textbf{74.8}\\
		& \textbf{CiT}  & ViT-H/14 & SWAG\cite{singh2022revisiting} & IG hashtags & 62 hrs & 
        \textbf{75.5}\\
		\hline
\end{tabular}
\vspace{-5pt}
\caption{Comparison to existing methods on YFCC and CC12M. 
Under identical vision encoders, CiT achieves +3.2\% higher accuracy with YFCC100M than using the human-cleaned YFCC15M subset and +5.9\% accuracy over LiT on YFCC15M.
* indicates reproduced results with  $\text{BERT}_\text{base}$ (uncased) for fair comparison;
see appendix for the implementation differences to original LiT~\cite{zhai2022lit}. Total time for training and curation is reported for 16 V100 GPUs and varies depending on quality of embeddings from the vision encoder.
}
\label{tbl:sota_yfcc}
\end{table*}

\begin{table*}[!h]
\centering
\tablestyle{4pt}{1.1}
\begin{tabular}{l l l c l l}
		Method & Vision Encoder & Vision Initialization & w/ Labeled Data & Total Time & 
        IN-1K Acc \\
		\shline
		\hline
		OpenCLIP & ViT-B/32 & scratch & \xmark & 458 hrs & 
        62.9\\
		OpenCLIP & ViT-B/16 & scratch & \xmark & 981 hrs & 
        67.1\\
		OpenCLIP & ViT-L/14 & scratch & \xmark & 6803 hrs & 
        72.8\\
		LiT\cite{zhai2022lit} & ViT-B/32 & AugReg\cite{steiner2021train} & IN-21K & 31 hrs & 
        62.8\\
		\hline
		\rowcolor{defaultcolor} \textbf{CiT}  & ViT-B/16 & MoCo-v3\cite{chen2021empirical} & \xmark & 26 hrs & 
        \textbf{67.1} \\
		\textbf{CiT}  & ViT-B/32 & AugReg\cite{steiner2021train} & IN-21K & 62 hrs & 
        \textbf{67.5} \\
		\textbf{CiT}  & ViT-B/16 & AugReg\cite{steiner2021train} & IN-21K & 63 hrs & 
        \textbf{73.1} \\
		\textbf{CiT}  & ViT-L/16 & AugReg\cite{steiner2021train} & IN-21K & 27 hrs & 
        \textbf{75.8} \\
		\textbf{CiT}  & ViT-H/14 & SWAG\cite{singh2022revisiting} & IG hashtags & 26 hrs & 
            \textbf{76.4} \\
		\hline
\end{tabular}
\vspace{-5pt}
\caption{CiT on LAION400M: CiT reaches OpenCLIP-level accuracy with 37$\times$ total training time improvement.} 
\vspace{-10pt}
\label{tbl:sota_laion}
\end{table*}

\subsection{Comparison to prior work on ImageNet}
We compare CiT with existing contrastive cross-modal models in Tables \ref{tbl:sota_yfcc} (YFCC and CC12M), \ref{tbl:sota_laion} (LAION400M) and \ref{tbl:sota_m2c2} (raw image-text crawl). We report the pre-training method (CLIP/LiT/CiT), vision encoder and initialization, usage of human-annotated labels, total training time in our setup (16 GPUs), as well as the ImageNet 0-shot accuracy.

\paragraph{YFCC.} In Table \ref{tbl:sota_yfcc} we report several data points for LiT and CiT training with various vision encoders and initialization.
On YFCC15M, CiT outperforms LiT on self-supervised MoCo-v3 vision encoders by +5.9\% accuracy.
With ViT-B/32 trained with supervised AugReg on IN-21K, CiT yields a +3.4\% gain over LiT.
On YFCC15M+CC12M data with ViT-L/16 models, CiT outperforms LiT by +3.4\%.

On YFCC100M we observe that LiT underperforms compared to YFCC15M (58.9 vs 59.9), due to cleaning~\cite{radford2021learning} of the 15M subset. CiT however can \textit{reverse} the trend. CiT outperforms its counterpart from YFCC15M by 3\%+ when using the less curated YFCC100M. 
This indicates human cleaning of YFCC100M by CLIP\cite{radford2021learning} is sub-optimal. The performance of CiT on YFCC100M is even \textbf{+2.6\%} better than LiT on YFCC15M+CC12M. This trend holds for larger image model sizes (ViT-L/H) and stronger initialization (AugReg/SWAG), which lead to better accuracy.

\paragraph{LAION400M.} In Table \ref{tbl:sota_laion}  we see that CiT performs better than OpenCLIP on LAION400M, while being substantially faster. For example, CiT with ViT-B/16 MoCo-v3 vision encoder performs as good as OpenCLIP but is 37.7\x faster in training. With more advanced initialization and larger ViT-L models, CiT is  283\x~faster and 3\% more accurate, producing 75.8\% in 1.1 days with a 16 GPU setup, while OpenCLIP would take \app283 days for an accuracy of 72.8\%. We note that this extreme speedup comes with the caveat that our approach curates data with respect to downstream tasks; therefore, 
CiT only uses 26 hours for training, compared to 981 hours for OpenCLIP pre-training.

\begin{table*}[!h]
\centering
\vspace{-5pt}
\tablestyle{4pt}{1.1}
\begin{tabular}{l l l c l l}
		Method & Vision Encoder & Vision Initialization & w/ Labeled Data & Total Time & 
        IN-1K Acc.\\
		\shline
		OpenCLIP & ViT-B/16 & from scratch & \xmark & n/a & 
        NaN loss \\
		LiT & ViT-B/16 & MoCo-v3\cite{chen2021empirical} & \xmark & n/a & 
        NaN loss \\
		LiT (English filter) & ViT-B/16 & MoCo-v3\cite{chen2021empirical} & \xmark & 65 hrs & 
        56.7\\
		\hline
		\rowcolor{defaultcolor} \textbf{CiT}  & ViT-B/16 & MoCo-v3\cite{chen2021empirical} & \xmark & 39 hrs & 
        \textbf{68.7}\\
		\textbf{CiT}  & ViT-B/32 & AugReg\cite{steiner2021train} & IN-21K & 69 hrs & 
        \textbf{68.4} \\
		\textbf{CiT}  & ViT-B/16 & AugReg\cite{steiner2021train} & IN-21K & 72 hrs & 
        \textbf{75.2} \\
		\textbf{CiT}  & ViT-L/16 & AugReg\cite{steiner2021train} & IN-21K & 105 hrs & 
        \textbf{77.9} \\
		\textbf{CiT}  & ViT-H/14 & SWAG\cite{singh2022revisiting} & IG hashtags & 43 hrs & 
        \textbf{77.4} \\
		\hline
\end{tabular}
\vspace{-5pt}
\caption{CiT on Raw Image-Text Crawl: CiT is able to produce strong results when learning from raw image-text data. The raw data contains 1.2B image-text pairs. An  English language filter, which reduces the data to 343M pairs, is required to stabilize LiT training.
}
\label{tbl:sota_m2c2}
\end{table*}

\begin{table*}[!t]
\vspace{-10pt}
\centering
\scalebox{0.7}{
\hspace{-5pt}
	\setlength\tabcolsep{2.7pt}
	\begin{tabular}{l  c | c c c c c c c c c c c c c c c c c c c c c c c c c c | c}
		 & Time 
		& \rotatebox[origin=l]{90}{Food-101} 
		& \rotatebox[origin=l]{90}{CIFAR10} 
		& \rotatebox[origin=l]{90}{CIFAR100} 
		& \rotatebox[origin=l]{90}{CUB}
		& \rotatebox[origin=l]{90}{SUN397}
		& \rotatebox[origin=l]{90}{Cars}
		& \rotatebox[origin=l]{90}{Aircraft}
		& \rotatebox[origin=l]{90}{DTD}
		& \rotatebox[origin=l]{90}{Pets}
		& \rotatebox[origin=l]{90}{Caltech-101}
		& \rotatebox[origin=l]{90}{Flowers}
		& \rotatebox[origin=l]{90}{MNIST}
		& \rotatebox[origin=l]{90}{FER-2013}
		& \rotatebox[origin=l]{90}{STL-10}
		& \rotatebox[origin=l]{90}{EuroSAT}
		& \rotatebox[origin=l]{90}{RESISC45}
		& \rotatebox[origin=l]{90}{GTSRB}
		& \rotatebox[origin=l]{90}{KITTI}
		& \rotatebox[origin=l]{90}{Country211}
		& \rotatebox[origin=l]{90}{PCAM}
		& \rotatebox[origin=l]{90}{UCF101}
		& \rotatebox[origin=l]{90}{Kinetics700}
		& \rotatebox[origin=l]{90}{CLEVR}
		& \rotatebox[origin=l]{90}{HatefulMemes}
		& \rotatebox[origin=l]{90}{SST2}
		& \rotatebox[origin=l]{90}{ImageNet}
		& \rotatebox[origin=l]{90}{Avg}\\
		\shline
		\hline
		CLIP~\cite{radford2021learning,mu2021slip} & 27 & 50.6 & 66.0 & 34.5 & 38.8 & 51.1 & 4.0 & 5.4 & 21.2 & 28.5 & 60.9 & 53.3 & 8.4 & 17.3 & 90.5 & 30.2 & 21.5 & 6.1 & 35.1 & 10.5 & 53.5 & 28.5 & 22.1 & 10.8 & 52.4 & 50.7 & 37.6 & 34.2\\
		SLIP~\cite{mu2021slip} & 41 & 59.5 & 78.6 & 45.2 & 38.7 & 53.4 & 5.4 & 5.7 & 26.1 & 31.1 & 71.0 & 56.6 & 9.8 & 19.6 & 94.4 & 20.3 & 28.9 & 14.5 & 34.0 & 11.6 & 55.4 & 37.7 & 26.9 & 17.5 & 52.8 & 51.1 & 42.8 & 38.0\\
		\shline
		{\small \textbf{CiT}-1K-meta} & 5 & 45.6 & 81.0 & 49.9 & 30.4 & 44.9 & 6.3 & 8.3 & 26.8 & \textbf{80.0} & 71.2 & 25.1 & 7.3 & 26.0 & 95.2 & 19.1 & 14.3 & 6.9 & 22.2 & 6.2 & \textbf{54.1} & 34.7 & 24.7 & 13.4 & 50.7 & 50.1 & \textbf{61.2} & 38.5\\
		{\small \textbf{CiT}-21K-meta} & 15  & 51.2 & \textbf{84.4} & 53.5 & 45.7 & \textbf{52.3} & 7.6 & 9.0 & 31.6 & 69.2 & 73.8 & 56.1 & 10.6 & 24.5 & 95.7 & \textbf{30.1} & \textbf{23.4} & \textbf{7.9} & 28.5 & 9.2 & 51.0 & 39.5 & \textbf{28.7} & \textbf{15.0} & 49.3 & 49.1 & 57.4 & 40.6\\
        {\small \textbf{CiT}-multi-meta} & 11  & 51.3 & 81.8 & 50.5 & 50.7 & 51.6 & 9.5 & \textbf{14.6} & 30.8 & 75.6 & 73.3 & 58.7 & 10.3 & \textbf{26.2} & 95.6 & 23.2 & 19.1 & 7.8 & 14.6 & \textbf{9.4} & 50.8 & 39.7 & 28.0 & 14.7 & \textbf{52.8} & 50.0 & 58.8 & 40.4\\
        {\small \textbf{CiT}-sep.-meta} & 7 & \textbf{59.1} & 82.2 & \textbf{55.2} & \textbf{56.6} & 50.7 & \textbf{13.0} & 13.1 & \textbf{32.8} & 74.8 & \textbf{77.6} & \textbf{65.9} & \textbf{16.9} & 13.8 & \textbf{96.3} & 17.1 & 21.6 & 7.6 & \textbf{40.6} & \textbf{9.4} & 53.5 & \textbf{42.7} & 27.8 & 14.2 & 52.2 & \textbf{50.9} & 50.7 & \textbf{42.2}\\
	\end{tabular}
}
\vspace{-10pt}
\caption{CiT on 26 zero-shot benchmarks when trained on YFCC15M. We vary metadata from IN-1K, IN-21K, combined (multi) as well as separate (sep.) across 26 tasks.
All methods use ViT-B/16 and we use MoCo-v3 vision initialization. Larger encoders are in Table~\ref{tbl:mt}. }
\label{tbl:compressed_mt}
\vspace{-10pt}
\end{table*}

\paragraph{Raw Image-Text Crawl.} 
We further test CiT on our raw image-text crawl containing 1.2B unfiltered image-text pairs from the web (about 343M pairs have English text). The data  contains a large degree of noise.
Results are shown in Table \ref{tbl:sota_m2c2}. 
To understand the challenge of training on raw image-text pairs, we run CLIP and LiT training on the raw image-text pairs. This yields 
unstable training that quickly reaches NaN loss for both a CLIP and LiT training. We believe some noisy pairs are unhealthy for training.
By using our English filter to clean the text, we can train LiT and it reaches 56.7\% IN-1K zero-shot accuracy. 
Training our CiT (\textit{without} even using an English filter) achieves 68.7\% which is \textbf{+12.0}\% higher.   
This indicates raw and very noisy image-text pairs lead to poor accuracy, but 
CiT can overcome this and curate high-quality data for vision-language learning.

Surprisingly, as shown in Table \ref{tbl:sota_laion}, CiT achieves much better performance than OpenCLIP trained on LAION400M. 
CiT on raw image-text reaches \textbf{77.9}\%, which is +5.1\% better than OpenCLIP ViT-L/14 (\cf Table \ref{tbl:sota_laion}). 
Note that our source is raw, 
with multilingual texts, whereas LAION400M is a curated English-only dataset filtered by the CLIP model. 
The training data used by CiT (\eg 131M for 77.9\%) is just around 1/5 of the scale of LAION400M dataset (one epoch), showing the effectiveness of curating training data.

\subsection{Comparison across 26 benchmarks}
We extend CiT to 26 common 0-shot evaluation tasks for CLIP/SLIP models~\cite{mu2021slip} on the public dataset YFCC15M.
We provide more comparisons with further encoders as well as pre-training on LAION400M in the appendix. We evaluate with prompts from CLIP/SLIP. For ImageNet, we drop the extra prompts used by LiT for a fair comparison with the baselines. 
We use three setups of metadata: (\textit{i}) IN-1K, (\textit{ii}) IN-21K, and (\textit{iii}) multi-task CiT that combines class names from all 26 tasks (\textit{iv}) we run every task separately on a \textit{single} GPU as a low-compute setup (this trains a model for each task with separate metadata). Results are in Table \ref{tbl:compressed_mt} and discussed next.  

We first evaluate CiT trained with IN-1K metadata on all 26 tasks. As expected accuracy on ImageNet and Pets is highest among the metadata variants (\textit{i-iv)}. Overall, we observe that \textit{CiT 1K meta} already exhibits certain generality to all tasks and can outperform CLIP (34.2 \vs 38.5\%) and is similar to SLIP, but 8.2\x~faster (5 \vs 41 hours), demonstrating its efficiency.

\begin{table*}[!t]
\centering
\vspace{-16pt}
\scalebox{0.75}{
\begin{tabular}{l|l|c|c}
Step ($c$) & Text & ImageNet Class & Cosine Sim. \\
\shline
0 & title: \textit{``Wollaston Beach''} & beach & 0.739 \\
100 & title: \textit{``tn\_kif\_3128''} & Vizsla & 0.779\\
1000 & tag: \textit{``beach plumisland parker river national wildlife refuge newburyport massachusetts ocean''} & beach & 0.716\\
2000 & desc: \textit{``These guys were nice, told me all about this and other planes of the show, but unfortunately...''} & military aircraft & 0.725\\
3000 & title: \textit{``Turtle''} & terrapin & 0.725\\
4000 & desc: \textit{``One of the fountains close by the south west entrance to the park''} & fountain & 0.734\\
5000 & title: \textit{``butterfly''} & Papillon & 0.735\\
5000 & tag: \begin{CJK}{UTF8}{min}\textit{``ash;explosion;sakurajima;kagoshima;桜島;鹿児島県;volcano;tarumizu;垂水市;japan;eruption;日本''}\end{CJK} & volcano & 0.645\\
\hline
\end{tabular}
}
\vspace{-10pt}
\caption{Samples of curated text over training steps ($c$) from YFCC100M. CiT uses MoCo-v3 initialized vision encoder. 
}
\label{tbl:curated_sample}
\vspace{-12pt}
\end{table*}

Next, we explore the WordNet lemma from ImageNet-21K as a relatively general metadata for training CiT.
In Table \ref{tbl:compressed_mt}, \textit{CiT-21K-meta} improves broadly over IN-1K leading to 40.6\% average accuracy, showing that a more general taxonomy works well across tasks.   

We combine the taxonomies from all 26 tasks in \textit{CiT-multi-meta}. This allows us to curate training data for all 26 tasks at again almost no extra training cost. We notice that multi-task CiT is on average similarly accurate as IN-21K metadata (40.4\% \vs 40.6\%) and converges faster because CiT is more targeted towards tasks of interest.

Finally, we compare a setup that trains a model for each task with separate metadata. \textit{CiT-sep.-meta} in Table \ref{tbl:compressed_mt} achieves overall the best average accuracy of 42.2\% across tasks. This setup uses a restricted 1-GPU setting to save compute and could be boosted further with longer training. We think that this scenario might be quite practical, where some domain data exists (\eg on bird images in CUB) and one wants to build a classification system given a large amount of noisy image-text data from the web. 

\subsection{Further Analysis}
\label{sec:discussion}

\paragraph{Samples of Curated Data}. 
We further investigate samples curated by CiT on YFCC100M dataset in Table \ref{tbl:curated_sample}. We show training steps, a sample text, the related ImageNet metadata, as well as the cosine similarity in CiT's data proxy. 
At step $c=0$ CiT's \textit{data proxy} tends to select text with similar length as class names and string-matching behavior; the short-term run of CiT (\eg $c=100$) has some matching issues with many false positives.
Later on, CiT starts to select texts of various lengths with similar semantics as the metadata. We do not observe any clearly less useful samples such as file names after $c=2000$. Interestingly, CiT can even use the English part of mixed language texts from YFCC100M (as in the last example).

\begin{figure}[h]
\vspace{-10pt}
\centering
\includegraphics[width=1\linewidth]{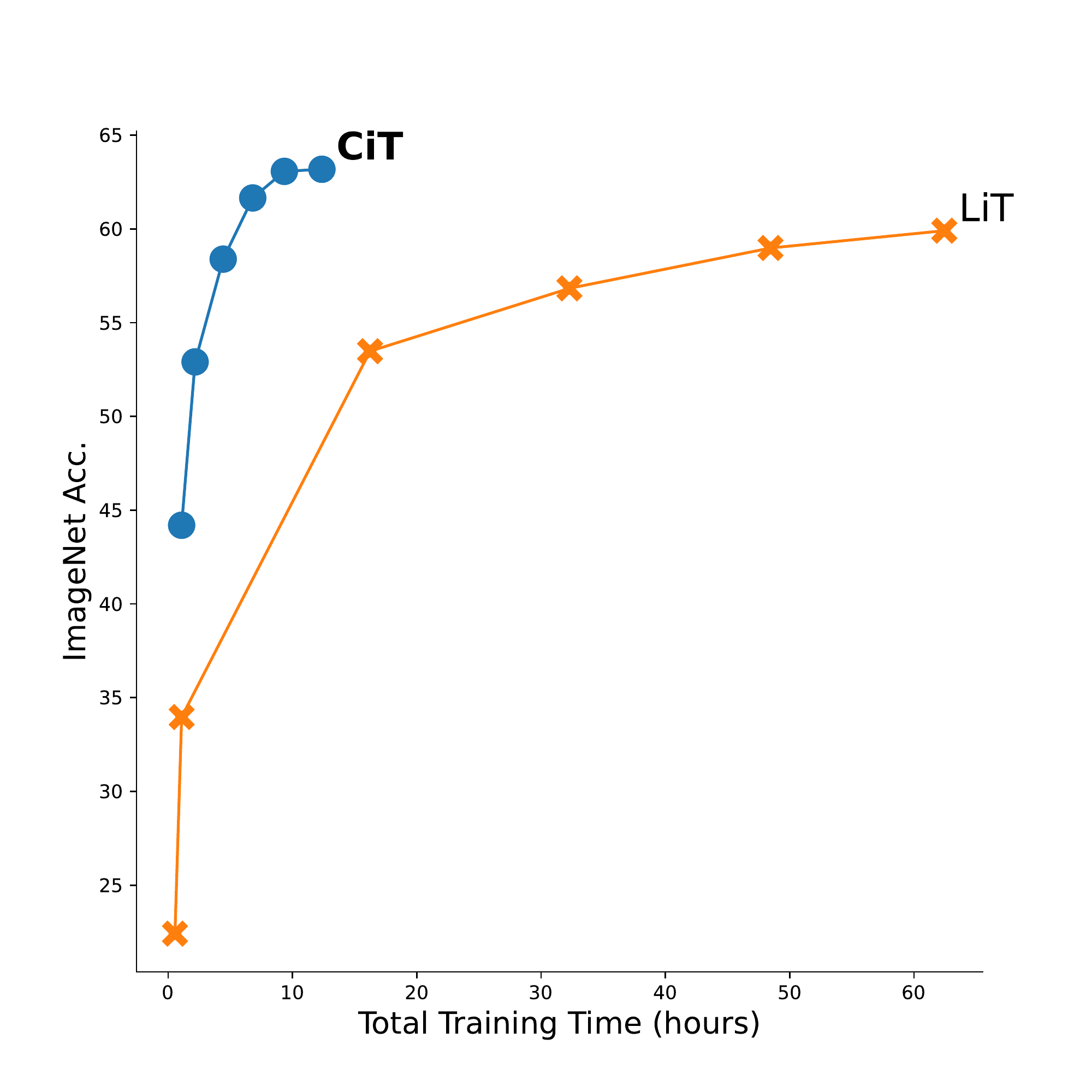}
\vspace{-20pt}
\caption{CiT on provides $>$5$\times$ speedup and +3.4\% accuracy gain over LiT\cite{zhai2022lit} on AugReg ViT-B/32 vision encoders. Training data is YFCC15M. 
Models are evaluated at 6 evenly sampled iterations.}
\label{fig:gain_time_lit}
\end{figure}

\paragraph{Speed/accuracy trade-off.} In Figure \ref{fig:gain_time_lit}, we show the speed/accuracy tradeoff of CiT \vs LiT~\cite{zhai2022lit}, corresponding to results in Table~\ref{tbl:sota_yfcc}). We see that CiT achieves a win-win scenario compared to LiT on identical AugReg ViT-B/32 vision encoders: a \textit{+3.4\%} higher accuracy on ImageNet, and a \textit{5$\times$} faster total training time (including the curation time).  
on data YFCC15M~\cite{radford2021learning}. 

\begin{figure}[t]
\centering
\vspace{-11pt}
\includegraphics[width=3.0in]{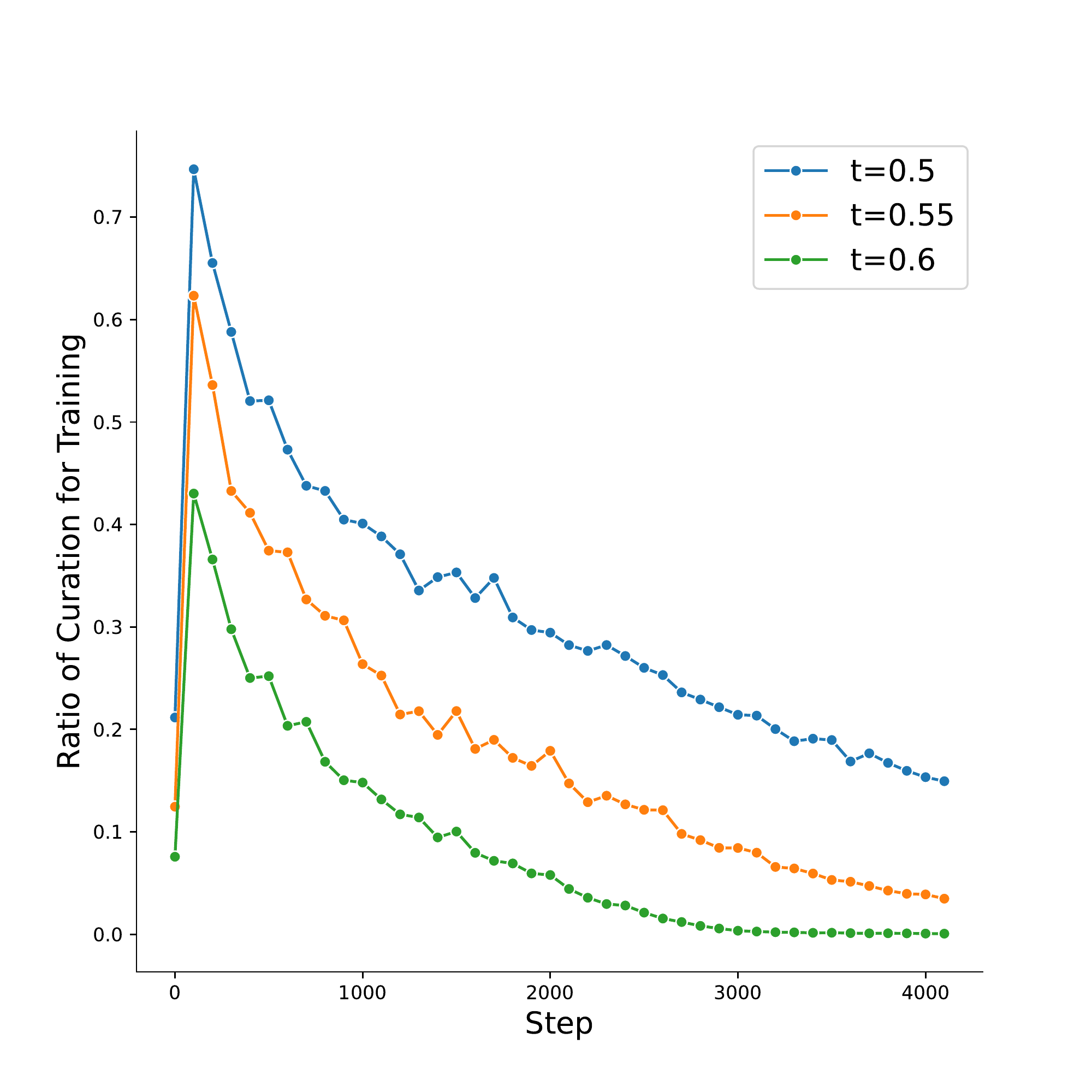}
\vspace{-20pt}
\caption{Ratio of curation under different thresholds $t$.
CiT broadly uses data first and curates more towards end of training.}
\vspace{-15pt}
\label{fig:ratio}
\end{figure}

\paragraph{Ratio of Curation}. We are interested in the training dynamics of CiT. We use different curation thresholds $t$ and inspect the amount of curated training data. In Figure \ref{fig:ratio}, we see that the ratio of curation which corresponds to the fraction of used training samples from the raw data source, see \S\ref{sec:curation_proxy}, keeps changing over steps for curation/training. Initially, CiT uses more data, \eg for a threshold of $t=0.5$, it peaks at about 75\%. In this phase, the latent space of the text encoder is less aligned with the vision latents. Later on during training, CiT 
starts to produce embeddings that better represent the downstream task, producing a lower ratio.

\section{Conclusion}
This paper contributes CiT, a novel learning algorithm for efficient pre-training from noisy image-text data.
CiT incorporates a curation process into learning to pull the training data distribution closer to downstream tasks. Our experiments demonstrate both significant accuracy and training time improvements when learning from either public or our own uncurated data from the web. We observe that training on the 
raw image-text pairs in YFCC can achieve better accuracy over the cleaned version from a hand-crafted filter pipeline.
Further, we show that CiT can train with raw image-text pairs crawled from the web, which would lead to instability for vanilla pre-training objectives. 

\paragraph{Acknowledgement.} We thank Norman Mu, Shang-Wen Li, Vasu Sharma, Wojciech Galuba and Max Bain for help.

\newpage
\appendix

\section{Appendix}
In this appendix, \S\ref{sec:impl} contains implementation details and \S\ref{sec:extra_results} contains further results as well as ablations.

\subsection{Implementation Details}
\label{sec:impl}

\subsubsection{PyTorch Pseudo Code} \label{sec:impl:pseudo}
To facilitate implementation of CiT, we provide the PyTorch pseudo-code in Algorithm \ref{alg:cit_pytorch} below.

\begin{algorithm}[!ht]
\DontPrintSemicolon
\caption{CiT: PyTorch Pseudo Code}
\label{alg:cit_pytorch}

\begin{minted}[fontsize=\footnotesize,xleftmargin=10pt,linenos]{python}
# b: maximum training steps as budget.
# d: raw data loader.
# t_meta: textual metadata.
# bsz: batch_size.
# t: threshold.
# gamma: target radio for curation.
# s: number of expected pairs.

c = 0
while c < b:
  if c % int(s//bsz) == 0:
    x_meta = model(t_meta)
    x_meta = normalize(x_meta)
    d_c = []
    while len(d_c) < s:
      x_imgs, x_txts = next(d)
      x_txts = model(x_txts)
      x_txts = normalize(x_txts)
      v = x_txts @ x_meta.t()
      sel = max(v) > t
      b_ratio = sum(sel) / len(sel)
      if b_ratio < gamma:
         sel = max(v).topk(
                k=int(bsz*gamma), dim=0)
      d_c.extend((x_imgs[sel], x_txts[sel]))
                
  for (x_imgs, x_txts) in batchify(d_c):
    x_imgs, x_txts = model(x_imgs, x_txts)
    x_imgs, x_txts = normalize(x_imgs,x_txts)
    # scale: learnable log logit scale
    l = exp(scale) * x_imgs @ x_txts.t()
    labels = arange(bsz)
    loss = cross_entropy(l, labels)
    loss.backward()
    c += 1
\end{minted}
\end{algorithm}

\subsubsection{Dataloader Implementation} For efficiency, we only load text during the curation loop and the training loop uses the curated indices to reload the full image-text pairs.
Our implementation also supports in-memory storage of curated image-text pairs in case the data source is not randomly accessible for (re-)loading curated data, where all $s$ pairs of training data can be stored in the CPU memory with  image tensors represented as \texttt{uint8} data.
We use a larger batch size for curation (compared to training) to speed up CiT. 

\subsubsection{Detailed Implementation Settings}
The hyper-parameters of CiT training are shown in Table \ref{tbl:hp}. We mostly follow~\cite{zhai2022lit,mu2021slip,radford2021learning}. CiT is trained on 16 GPUs with a global batch size of 16,384 (1024 per GPU). 

Hyperparameters for CiT curation outlined in \S3 of the main paper are shown in Table \ref{tbl:hp_curation}.
We use different thresholds $t$ and minimal ratios $\gamma$ for each dataset/metadata combination to fit the training into a budget $b$ shown in the table as well. 
We use the same values for all variants of vision encoders. Due to smaller size, we use a lower $t$ for YFCC15M and CC12M, whereas for YFCC100M and Raw Img-Text Crawl we use a higher $t$ to focus on high-quality data from the raw data source, in order to roughly meet the budget $b$.

\begin{table}[t]\centering%
\scalebox{0.92}{
\tablestyle{2pt}{1.1}
\setlength\tabcolsep{3.0pt}
\begin{tabular}{l|c}
	Hyperparameter & Value \\ 
	\toprule 
	Optimizer & AdamW \\
	Optimizer momentum & $\beta_1=0.9$, $\beta_2=0.999$ \\
	Optimizer $\epsilon$ & 1e-8\\
	Weight Decay (proj.) & 1.0\\
	Weight Decay (other) & 0.2 \\
	Base Learning Rate & 5e-4\\
	Learning Rate Schedule & cosine decay\\
	Minimum Learning Rate & 1e-5\\
	Gradient Clipping & None\\
	Warm-up \% of Train Steps & 4\% \\
	Batch size & 16,384 \\
	GPUs & 16 Nvidia V100 32GB GPUs \\
	precision & float16 \\
	Max BERT len. & 32\\
	Train Aug. & RandomResizedCrop(224, scale=(0.5, 1.0))\\
	YFCC15M/YFCC100M Aug. & shuffle/join tags\cite{zhai2022lit}\\
	Eval Aug. & Resize(256), CenterCrop(224)\\
	AugReg rgb Mean & (0.5, 0.5, 0.5)\\ 
	AugReg rgb Std. & (0.5, 0.5, 0.5)\\ 
	Other encoder rgb Mean & (0.485, 0.456, 0.406)\\ 
	Other encoder rgb Std. & (0.229, 0.224, 0.225)\\
\end{tabular}
}
\vspace{-5pt}
\caption{Hyperparameters of CiT Training.}
\label{tbl:hp}
\end{table}

\begin{table}[t]\centering%
\scalebox{0.9}{
\tablestyle{2pt}{1.1}
\setlength\tabcolsep{3.0pt}
\begin{tabular}{l l|c c c }
	Data Source & Metadata & $b$ & $t$  & $\gamma$   \\
	\toprule
	\textsc{YFCC15M} & IN-1K & 5K & 0.55 & 0.003 \\ 
	\textsc{YFCC15M} & IN-21K & 8K & 0.55 & 0.003 \\
	\textsc{YFCC15M} & multi. & 8K & 0.55 & 0.003 \\

	\textsc{YFCC100M} & IN-1K & 5K & 0.7 & 0.01 \\
	\textsc{LAION400M} & IN-1K & 5K & 0.6 & 0.01 \\
	\textsc{LAION400M} & IN-21K & 30K & 0.65 & 0.01 \\
	\textsc{LAION400M} & multi. & 16K & 0.6 & 0.01 \\
	\textsc{Raw Img-Txt} & IN-1K & 8K & 0.7 & 0.003 \\
	\textsc{Raw Img-Txt} & IN-21K & 60K & 0.75 & 0.003\\
	\textsc{Raw Img-Txt} & multi. & 30K & 0.7 & 0.003\\
\end{tabular}
}\vspace{-5pt}
\caption{Hyperparameters of CiT Curation. }
\label{tbl:hp_curation}
\end{table}

\paragraph{Single GPU Setting.}
\label{sec:single_gpu}
We provide more details on the implementation of the extremely efficient single GPU setup used for zero-shot evaluation on multiple tasks in Table \ref{tbl:mt}.
We can fit a batch size of 1,536 into a \textit{single} 32GB V100 GPU and train for $b=5000$ steps.
To ensure the training can be finished quickly, we set $\gamma=0.05$. 
Further to reduce the chance of using the minimal ratio during curation, we perform a pre-curation on YFCC15M for each task using BERT-SimCSE with a threshold of 0.45 to remove pairs with low relevance. 

\subsubsection{Implementation Differences from LiT}
\label{sec:littrick}

While we aim for a close reproduction of LiT~\cite{zhai2022lit}, there are a few tricks that our implementation does not incorporate and we suspect the differences on our LiT reproduction on YFCC stem from those.
Below we list some tricks known to us, but there could be more differences we are not aware of since we have no access to LiT's full preprocessing and training code.

\paragraph{Preprocessing.} For the captions, LiT performs extra filtering and removes titles that start with ``DSC'', ``IMG'', ``Picture''. Also, LiT removes text consisting of only the word ``image'' or text that contains a large fraction of digits.

\paragraph{Joint Contrastive Loss.} LiT adopts a joint contrastive loss over 3 text fields in YFCC15M and shows the gain in Figure 8 of the LiT paper~\cite{zhai2022lit}. Since this technique is specific to the type of captions in the specific YFCC data, we remove it from our implementation and randomly sample one of the three text fields to pair with a training image.

\paragraph{Text encoder.}
LiT adopts various text encoders such as $\text{BERT}_\text{base}$ and $\text{BERT}_\text{large}$. This work consistently uses $\text{BERT}_\text{base}$ for all main results to have a fair comparison.

\subsection{Additional Results} \label{sec:extra_results} 
This section extends the results of CiT in the main paper to full results across 26 CLIP/SLIP benchmarks on YFCC15M and LAION400M and an extra ablation study.

\subsubsection{Full Results on YFCC15M}

We show the full results of Table~\ref{tbl:compressed_mt} above in Table \ref{tbl:mt} below.
On average, CiT-multi-meta (52.6) is slightly better than CiT-21K-meta (51.7), which is better than CiT-sep-meta and CiT-1K-meta (47.2). It appears that the  broader ImageNet-21K wordnet taxonomy works well across datasets, and combining metadata from all downstream tasks is only slightly better than that.
We note that training on the larger metadata does not introduce much extra curation compute since forwarding the raw examples takes the majority of computation.
Nevertheless, we observe that larger metadata takes longer to converge and therefore increase the training budget to $b=8000$ for CiT-21K-meta and CiT-multi-meta. We expect larger budgets will lead to even better results.

Besides what was already discussed in the main paper, we observe that CiT performs even better on larger models or models trained with supervised (AugReg IN-21K) or weakly supervised (SWAG) data than the unsupervisedly pre-trained MoCo-v3 on IN-1K. Out-of-domain issues (\eg MNIST) are present even for larger vision encoders.

\begin{table}[t!]\centering
	\vspace{10pt}
	\tablestyle{5pt}{1.1}
	
	\subfloat[{Evaluation Prompts}]{
		\begin{tabular}{l r}
			Eval. Prompts & Acc\\
			\toprule
			\rowcolor{defaultcolor} CLIP+LiT prompts & \textbf{61.4} \\
			CLIP prompts only & 61.2 \\

		\end{tabular}
		\label{tbl:ablation:tt}
	}
	\hspace{5mm}
	\subfloat[{Training Objective}]{
		\begin{tabular}{l r}
			
			Objective
			& Acc\\
			\toprule
			\rowcolor{defaultcolor} img2txt obj. & \textbf{61.4} \\
			CLIP obj. & 61.2 \\
			
		\end{tabular}
		\label{tbl:ablation:objective}
	}
	\vspace{-5pt}
	\caption{\textbf{Additional ablation experiments}. We use the default setup (MoCo-v3 / $\text{BERT}_\text{base}$-SimCSE) and YFCC15M as data source and report IN-1K Accuracy.}
	\label{tbl:more_ablation}
	\vspace{-1.em}
\end{table}

\begin{table*}[!t]
    \vspace{10pt}
	\centering
	\scalebox{0.66}{
		\setlength\tabcolsep{2.7pt}
  \begin{tabular}{l l c | c c c c c c c c c c c c c c c c c c c c c c c c c c | c}
			Vis. Encoder & Init. & Hrs 
			& \rotatebox[origin=l]{90}{Food-101} 
			& \rotatebox[origin=l]{90}{CIFAR10} 
			& \rotatebox[origin=l]{90}{CIFAR100} 
			& \rotatebox[origin=l]{90}{CUB}
			& \rotatebox[origin=l]{90}{SUN397}
			& \rotatebox[origin=l]{90}{Cars}
			& \rotatebox[origin=l]{90}{Aircraft}
			& \rotatebox[origin=l]{90}{DTD}
			& \rotatebox[origin=l]{90}{Pets}
			& \rotatebox[origin=l]{90}{Caltech-101}
			& \rotatebox[origin=l]{90}{Flowers}
			& \rotatebox[origin=l]{90}{MNIST}
			& \rotatebox[origin=l]{90}{FER-2013}
			& \rotatebox[origin=l]{90}{STL-10}
			& \rotatebox[origin=l]{90}{EuroSAT}
			& \rotatebox[origin=l]{90}{RESISC45}
			& \rotatebox[origin=l]{90}{GTSRB}
			& \rotatebox[origin=l]{90}{KITTI}
			& \rotatebox[origin=l]{90}{Country211}
			& \rotatebox[origin=l]{90}{PCAM}
			& \rotatebox[origin=l]{90}{UCF101}
			& \rotatebox[origin=l]{90}{Kinetics700}
			& \rotatebox[origin=l]{90}{CLEVR}
			& \rotatebox[origin=l]{90}{HatefulMemes}
			& \rotatebox[origin=l]{90}{SST2}
			& \rotatebox[origin=l]{90}{ImageNet}
			& \rotatebox[origin=l]{90}{Avg}\\
			\shline
			\hline
			\multicolumn{2}{l}{\textsc{CLIP} \cite{radford2021learning,mu2021slip}} \\
			ViT-B/16 & scratch & 27 & 
                50.6 & 66.0 & 34.5 & 38.8 & 51.1 & 4.0 & 5.4 & 21.2 & 28.5 & 60.9 & 53.3 & 8.4 & 17.3 & 90.5 & 30.2 & 21.5 & 6.1 & 35.1 & 10.5 & 53.5 & 28.5 & 22.1 & 10.8 & 52.4 & 50.7 & 37.6 & 34.2\\
			ViT-L/16 & scratch & 189 & 
                59.5 & 72.9 & 41.5 & 40.3 & 53.6 & 6.9 & 6.4 & 20.6 & 27.9 & 65.4 & 55.0 & 10.3 & 34.5 & 94.2 & 22.7 & 28.8 & 5.8 & 41.4 & 12.6 & 54.9 & 34.3 & 24.0 & 12.9 & 54.3 & 50.1 & 40.4 & 37.4\\
			\hline
			SLIP\cite{mu2021slip}\\
			ViT-B/16 & scratch & 41 & 
                59.5 & 78.6 & 45.2 & 38.7 & 53.4 & 5.4 & 5.7 & 26.1 & 31.1 & 71.0 & 56.6 & 9.8 & 19.6 & 94.4 & 20.3 & 28.9 & 14.5 & 34.0 & 11.6 & 55.4 & 37.7 & 26.9 & 17.5 & 52.8 & 51.1 & 42.8 & 38.0\\
			ViT-L/16 & scratch & 284 & 
                64.4 & 87.8 & 56.4 & 39.8 & 58.9 & 8.6 & 7.8 & 26.8 & 32.0 & 76.6 & 59.4 & 13.2 & 36.0 & 96.6 & 27.7 & 36.5 & 7.2 & 28.8 & 15.6 & 54.4 & 42.6 & 30.0 & 14.1 & 53.4 & 50.1 & 46.2 & 41.2\\
			\shline
			\multicolumn{2}{l}{\textbf{CiT}-1K-meta} \\
			\rowcolor{defaultcolor} ViT-B/16 & MoCo-v3 & 5 & 
            45.6 & 81.0 & 49.9 & 30.4 & 44.9 & 6.3 & 8.3 & 26.8 & 80.0 & 71.2 & 25.1 & 7.3 & 26.0 & 95.2 & 19.1 & 14.3 & 6.9 & 22.2 & 6.2 & 54.1 & 34.7 & 24.7 & \textbf{13.4} & 50.7 & 50.1 & 61.2 & 38.5\\
			ViT-B/16 & AugReg & 8 & 
                57.9 & 92.3 & 74.2 & \textbf{36.9} & 52.5 & 7.7 & 5.6 & 25.2 & 77.9 & 84.5 & 38.8 & 8.3 & 31.2 & 94.4 & 16.6 & 24.3 & 6.5 & 17.2 & 6.4 & \textbf{59.1} & 47.8 & 32.2 & 13.3 & 52.0 & 50.1 & 68.9 & 41.6\\
			ViT-L/16 & AugReg & 8 & 
                60.0 & \textbf{93.6} & \textbf{77.8} & 36.3 & 54.0 & 9.0 & 5.7 & 25.6 & 79.8 & \textbf{87.3} & 45.2 & \textbf{9.7} & 29.2 & 96.1 & 20.9 & 32.8 & 7.0 & \textbf{36.0} & 7.6 & 52.8 & 51.5 & 35.2 & 12.6 & \textbf{53.0} & 49.7 & 71.6 & 43.8\\
			ViT-H/14 & SWAG & 11 & 
                \textbf{79.0} & 91.6 & 68.1 & 35.3 & \textbf{56.9} & \textbf{26.2} & \textbf{12.5} & \textbf{30.0} & \textbf{88.8} & 86.4 & \textbf{47.6} & 8.1 & \textbf{31.3} & \textbf{97.8} & \textbf{27.6} & \textbf{46.4} & \textbf{7.3} & 34.2 & \textbf{14.5} & 50.3 & \textbf{54.7} & \textbf{43.8} & 12.3 & 51.8 & \textbf{51.0} & \textbf{73.3} & \textbf{47.2}\\
			\hline
			\multicolumn{2}{l}{\textbf{CiT}-21K-meta} \\
			\rowcolor{defaultcolor} ViT-B/16 & MoCo-v3 & 15 & 
            51.2 & 84.4 & 53.5 & 45.7 & 52.3 & 7.6 & 9.0 & 31.6 & 69.2 & 73.8 & 56.1 & 10.6 & 24.5 & 95.7 & 30.1 & 23.4 & 7.9 & \textbf{28.5} & 9.2 & 51.0 & 39.5 & 28.7 & 15.0 & 49.3 & 49.1 & 57.4 & 40.6\\
			ViT-B/16 & AugReg & 23 & 
                75.3 & 93.8 & 75.7 & 57.8 & 59.8 & 9.7 & 10.1 & 35.4 & 68.3 & 87.9 & 74.3 & 12.1 & 27.4 & 97.1 & 30.8 & 30.6 & 7.3 & 24.3 & 9.9 & 50.5 & 54.7 & 37.4 & 13.6 & \textbf{53.8} & \textbf{50.1} & 63.7 & 46.6\\
			ViT-L/16 & AugReg & 29 & 
                78.9 & \textbf{95.1} & \textbf{78.6} & \textbf{60.5} & 61.9 & 11.6 & 10.9 & 35.1 & 74.2 & 90.5 & 75.4 & \textbf{14.8} & \textbf{34.8} & 98.0 & 24.7 & 35.5 & 7.5 & 25.7 & 10.9 & \textbf{50.8} & 57.4 & 40.7 & \textbf{14.8} & 49.9 & 48.7 & 67.7 & 48.3\\
			ViT-H/14 & SWAG & 39 & 
                \textbf{92.2} & 92.9 & 70.9 & 59.0 & \textbf{64.7} & \textbf{36.9} & \textbf{14.9} & \textbf{40.3} & \textbf{87.7} & \textbf{90.9} & \textbf{77.4} & 10.1 & 32.7 & \textbf{99.1} & \textbf{38.8} & \textbf{53.2} & \textbf{9.3} & 15.9 & \textbf{20.5} & 50.7 & \textbf{62.2} & \textbf{49.4} & 12.9 & 46.8 & 44.2 & \textbf{71.4} & \textbf{51.7}\\
			\hline
			\multicolumn{2}{l}{\textbf{CiT}-multi-meta} \\
			\rowcolor{defaultcolor} ViT-B/16 & MoCo-v3 & 11 & 
            51.3 & 81.8 & 50.5 & 50.7 & 51.6 & 9.5 & 14.6 & 30.8 & 75.6 & 73.3 & 58.7 & 10.3 & 26.2 & 95.6 & 23.2 & 19.1 & 7.8 & 14.6 & 9.4 & 50.8 & 39.7 & 28.0 & 14.7 & 52.8 & 50.0 & 58.8 & 40.4\\
			ViT-B/16 & AugReg & 11 & 
                77.8 & 94.0 & 76.5 & 63.9 & 60.1 & 10.3 & 13.1 & 35.2 & 79.0 & 88.9 & 79.4 & 12.2 & 33.0 & 96.2 & \textbf{31.6} & 29.3 & 10.2 & \textbf{17.4} & 9.6 & 50.8 & 56.0 & 38.0 & 12.5 & \textbf{55.8} & 47.8 & 67.0 & 47.9\\
			ViT-L/16 & AugReg & 16 & 
                80.4 & \textbf{95.3} & \textbf{79.4} & 65.6 & 61.9 & 13.3 & 11.3 & 35.1 & 79.9 & 90.6 & 80.1 & 10.7 & \textbf{37.8} & 97.4 & 29.3 & 35.0 & 7.8 & 13.8 & 10.7 & 49.7 & 59.5 & 41.3 & 13.0 & 54.5 & \textbf{47.9} & 70.5 & 48.9\\
			ViT-H/14 & SWAG & 31 & 
                \textbf{91.8} & 90.7 & 71.3 & \textbf{65.6} & \textbf{62.4} & \textbf{47.9} & \textbf{19.7} & \textbf{40.8} & \textbf{91.7} & \textbf{91.3} & \textbf{81.2} & \textbf{10.7} & 37.5 & \textbf{98.0} & 23.9 & \textbf{46.4} & \textbf{11.0} & 12.4 & \textbf{20.2} & \textbf{51.3} & \textbf{64.3} & \textbf{50.2} & \textbf{13.5} & 54.6 & 47.1 & \textbf{73.4} & \textbf{52.6}\\
			\hline\hline
			\multicolumn{4}{l}{\textbf{CiT}-sep.-meta (single GPU)} \\
			\rowcolor{defaultcolor} ViT-B/16 & MoCo-v3 & 4 & 
            59.1 & 82.2 & 55.2 & 56.6 & 50.7 & 13.0 & 13.1 & 32.8 & 74.8 & 77.6 & 65.9 & 16.9 & 13.8 & 96.3 & 17.1 & 21.6 & 7.6 & \textbf{40.6} & 9.4 & 53.5 & 42.7 & 27.8 & 14.2 & 52.2 & 50.9 & 50.7 & 42.2\\
			ViT-B/16 & AugReg & 5 & 
            79.1 & 94.4 & 75.2 & 73.8 & \textbf{60.6} & 19.4 & \textbf{17.4} & 36.6 & 78.1 & 88.0 & 79.8 & 12.4 & \textbf{39.2} & \textbf{97.0} & \textbf{31.1} & 29.1 & 11.1 & 30.1 & 9.9 & 51.9 & 54.9 & 37.1 & \textbf{19.2} & \textbf{52.5} & 50.0 & 56.8 & 49.4\\
			ViT-L/16 & AugReg & 7 & 
            83.8 & \textbf{94.8} & \textbf{79.6} & \textbf{76.9} & 60.4 & 19.6 & 17.2 & 36.0 & 77.8 & 89.6 & 82.2 & 12.1 & 39.0 & 96.7 & 24.8 & \textbf{31.2} & 9.7 & 26.9 & 10.7 & 57.6 & 59.1 & 39.9 & 14.9 & 46.8 & 51.2 & 60.1 & \textbf{49.9}\\
			ViT-H/14 & SWAG & 11 & 
            \textbf{92.1} & 89.9 & 71.8 & 71.3 & \textbf{65.4} & 52.0 & \textbf{20.9} & \textbf{38.7} & \textbf{90.6} & \textbf{90.4} & \textbf{84.8} & \textbf{15.1} & 30.6 & 92.8 & 26.8 & \textbf{47.1} & \textbf{13.4} & 34.8 & \textbf{20.8} & \textbf{59.4} & \textbf{65.8} & \textbf{50.1} & 14.0 & 48.5 & \textbf{51.7} & \textbf{67.0} & \textbf{54.1}\\
		\end{tabular}
	}
	\caption{CiT trained on YFCC15M and evaluated on 26 CLIP/SLIP benchmarks: we vary metadata on IN-1K, IN-21K and combined class names on 26 tasks (CiT-multi-meta) with a single training and run 26 separate training on each task with a single GPU (CiT-sep.-meta). }
\label{tbl:mt}
\end{table*}

\subsubsection{Full Results on LAION400M}

In Table \ref{tbl:mt_laion}, we show the result of CiT trained on LAION400M and evaluated on 26 CLIP/SLIP benchmarks.
With a larger data source, we realize CiT takes more time to converge especially with more metadata, which can be attributed to more data meeting the curation criteria. We set $b=16000$ for CiT-multi-meta and $b=30000$ for CiT-21K-meta.
The trend is similar to YFCC15M but with better performance aross the benchmarks.
Similar as in Table \ref{tbl:mt}, CiT-multi-meta is better than CiT-21K-meta, but this time the gap is larger. In addition to the longer training, we believe that the combined metadata from 26 benchmarks are more effective on larger pre-training data. 

\begin{table*}[!htp]
\centering
\vspace{-5pt}
\scalebox{0.66}{
	\setlength\tabcolsep{2.7pt}
 \begin{tabular}{l l c | c c c c c c c c c c c c c c c c c c c c c c c c c c | c}
		Vis. Encoder & Init. & Hrs 
		& \rotatebox[origin=l]{90}{Food-101} 
		& \rotatebox[origin=l]{90}{CIFAR10} 
		& \rotatebox[origin=l]{90}{CIFAR100} 
		& \rotatebox[origin=l]{90}{CUB}
		& \rotatebox[origin=l]{90}{SUN397}
		& \rotatebox[origin=l]{90}{Cars}
		& \rotatebox[origin=l]{90}{Aircraft}
		& \rotatebox[origin=l]{90}{DTD}
		& \rotatebox[origin=l]{90}{Pets}
		& \rotatebox[origin=l]{90}{Caltech-101}
		& \rotatebox[origin=l]{90}{Flowers}
		& \rotatebox[origin=l]{90}{MNIST}
		& \rotatebox[origin=l]{90}{FER-2013}
		& \rotatebox[origin=l]{90}{STL-10}
		& \rotatebox[origin=l]{90}{EuroSAT}
		& \rotatebox[origin=l]{90}{RESISC45}
		& \rotatebox[origin=l]{90}{GTSRB}
		& \rotatebox[origin=l]{90}{KITTI}
		& \rotatebox[origin=l]{90}{Country211}
		& \rotatebox[origin=l]{90}{PCAM}
		& \rotatebox[origin=l]{90}{UCF101}
		& \rotatebox[origin=l]{90}{Kinetics700}
		& \rotatebox[origin=l]{90}{CLEVR}
		& \rotatebox[origin=l]{90}{HatefulMemes}
		& \rotatebox[origin=l]{90}{SST2}
		& \rotatebox[origin=l]{90}{ImageNet}
		& \rotatebox[origin=l]{90}{Avg}\\
		\shline
		\hline
		\multicolumn{3}{l}{CLIP (WIT400M) \cite{radford2021learning}} \\
		ViT-B/32 & scratch & 458 & 
            84.4 & 91.3 & 65.1 & 37.8 & 63.2 & 59.4 & 21.2 & 44.5 & 87.0 & 87.9 & 66.7 & 51.9 & 47.3 & 97.2 & 49.4 & 60.3 & 32.2 & 39.4 & 17.8 & 58.4 & 64.5 & 47.8 & 24.8 & 57.6 & 59.6 & 63.2 & 56.9\\
		ViT-B/16 & scratch& 981 & 
            89.2 & 91.6 & 68.7 & 39.1 & 65.2 & 65.6 & 27.1 & 46.0 & 88.9 & 89.3 & 70.4 & 56.0 & 52.7 & 98.2 & 54.1 & 65.5 & 43.3 & 44.0 & 23.3 & 48.1 & 69.8 & 52.4 & 23.4 & 61.7 & 59.8 & 68.6 & 60.1\\
		ViT-L/14 & scratch & 6803 & 
            92.9 & 96.2 & 77.9 & 48.3 & 67.7 & 77.3 & 36.1 & 55.3 & 93.5 & 92.6 & 78.7 & 87.2 & 57.5 & 99.3 & 59.9 & 71.6 & 50.3 & 23.1 & 32.7 & 58.8 & 76.2 & 60.3 & 24.3 & 63.3 & 64.0 & 75.3 & 66.2\\
		\hline
		\multicolumn{3}{l}{OpenCLIP *} \\
		ViT-B-32 & scratch & 458 & 
            n/a & 90.8 & 70.2 & n/a & 67.0 & 79.2 & 16.8 & 54.3 & 86.8 & 83.3 & 68.3 & 37.4 & 42.7 & 95.5 & 51.6 & n/a & 42.0 & 28.8 & 14.7 & 54.6 & n/a & n/a & 16.3 & n/a & 52.6 & 62.9 & n/a\\
		ViT-B-16 & scratch & 981 & 
            n/a & 91.7 & 71.0 & n/a & 69.6 & 83.7 & 17.5 & 51.3 & 89.2 & 83.5 & 69.3 & 66.6 & 42.9 & 97.0 & 50.3 & n/a & 43.5 & 19.0 & 18.1 & 60.5 & n/a & n/a & 28.8 & n/a & 54.7 & 67.0 & n/a\\
		ViT-L-14 & scratch & 6803 & 
            n/a & 94.7 & 77.4 & n/a & 72.6 & 89.6 & 25.1 & 60.3 & 91.9 & 84.2 & 75.4 & 76.4 & 50.1 & 98.0 & 61.8 & n/a & 50.0 & 20.8 & 23.1 & 48.6 & n/a & n/a & 24.2 & n/a & 56.3 & 72.7 & n/a\\
		\shline
		\multicolumn{3}{l}{\textbf{CiT}-1K-meta} \\
		\rowcolor{defaultcolor} ViT-B/16 & MoCo-v3 & 26 & 
        31.2 & 80.7 & 56.7 & 29.5 & 41.7 & 12.6 & 3.9 & 35.2 & 85.9 & 82.3 & 19.1 & 16.3 & 25.0 & 89.7 & 20.0 & 19.7 & \textbf{14.5} & 42.2 & 3.7 & 55.3 & 34.8 & 23.0 & 14.4 & 49.5 & 49.3 & 67.0 & 38.6\\
		ViT-B/32 & AugReg & 62 & 
        45.0 & 86.6 & 68.8 & 34.5 & 48.1 & 12.1 & 3.8 & 35.3 & 87.0 & 87.6 & 34.5 & 10.2 & 29.2 & 89.8 & 19.7 & 23.0 & 10.5 & 33.1 & 4.4 & 50.6 & 45.5 & 27.7 & 15.2 & 48.5 & \textbf{50.4} & 67.5 & 41.1\\
            ViT-B/16 & AugReg & 63 &
            45.4 & 87.8 & 70.9 & 33.7 & 50.8 & 12.4 & 3.3 & 38.0 & 86.2 & 89.0 & 31.5 & 9.7 & 26.4 & 90.0 & 25.3 & 25.3 & 13.2 & \textbf{34.9} & 5.2 & \textbf{54.7} & 50.0 & 31.5 & 14.7 & 50.4 & 49.3 & 73.0 & 42.4\\
		ViT-L/16 & AugReg & 27 & 
        45.3 & \textbf{90.6} & \textbf{76.3} & 36.3 & 54.7 & 13.6 & 5.0 & 35.9 & 87.2 & \textbf{92.1} & 32.0 & 10.2 & 20.0 & 91.3 & 28.2 & 31.2 & 10.6 & 21.4 & 5.5 & 51.7 & 50.9 & 33.6 & \textbf{16.1} & 48.9 & 50.1 & 75.7 & 42.9\\
		ViT-H/14 & SWAG & 26 & 
        \textbf{65.4} & 89.8 & 68.7 & \textbf{36.4} & \textbf{56.5} & \textbf{38.0} & \textbf{7.9} & \textbf{41.7} & \textbf{89.4} & 88.5 & \textbf{41.4} & \textbf{10.2} & \textbf{30.5} & \textbf{94.3} & \textbf{34.6} & \textbf{41.5} & 12.0 & 19.1 & \textbf{12.3} & 49.5 & \textbf{57.0} & \textbf{42.6} & 13.2 & \textbf{51.5} & 46.5 & \textbf{76.2} & \textbf{46.7}\\
		\hline
		\multicolumn{3}{l}{\textbf{CiT}-21K-meta} \\
		\rowcolor{defaultcolor} ViT-B/16 & MoCo-v3 & 70 & 
        64.8 & 85.0 & 63.1 & 59.5 & 56.3 & 26.2 & 8.1 & 40.2 & 87.6 & 87.1 & 60.6 & 17.8 & 34.5 & 95.9 & 29.4 & 30.3 & 10.9 & 33.0 & 6.4 & \textbf{54.5} & 48.8 & 31.2 & 15.1 & 47.9 & 50.1 & 64.1 & 46.5\\
		ViT-B/32 & AugReg & 57 & 
        71.7 & 91.1 & 72.8 & 62.4 & 59.0 & 18.8 & 5.9 & 42.6 & 81.8 & 89.8 & 67.5 & 16.3 & 38.8 & 96.3 & 27.1 & 32.8 & 12.4 & 33.9 & 6.4 & 52.8 & 56.8 & 35.9 & \textbf{16.4} & 51.0 & 50.1 & 65.0 & 48.3\\
		ViT-B/16 & AugReg & 72 & 
        77.1 & 92.8 & 74.7 & \textbf{68.9} & 61.9 & 20.6 & 8.3 & 41.5 & 85.7 & 91.2 & 73.8 & \textbf{21.7} & 38.3 & 97.0 & 26.2 & 36.4 & \textbf{15.1} & \textbf{41.8} & 7.1 & 52.4 & 56.8 & 38.3 & 12.1 & 51.0 & \textbf{50.5} & 71.2 & 50.5\\
		ViT-L/16 & AugReg & 97 & 
        77.5 & \textbf{93.5} & \textbf{79.1} & 67.6 & 62.9 & 19.5 & 8.3 & 44.8 & 84.4 & 93.1 & 71.5 & 18.9 & 34.2 & \textbf{98.0} & \textbf{29.6} & 38.9 & 11.7 & 22.9 & 7.7 & 50.9 & 60.3 & 41.6 & 14.8 & 51.5 & 48.2 & 73.9 & 50.2\\
		ViT-H/14 & SWAG & 135 & 
        \textbf{89.2} & 91.5 & 72.1 & 68.2 & \textbf{64.0} & \textbf{36.9} & \textbf{10.4} & \textbf{43.9} & 88.2 & \textbf{92.1} & \textbf{75.8} & 7.1 & \textbf{41.7} & 97.4 & 29.2 & \textbf{49.6} & 10.7 & 34.6 & \textbf{15.0} & 50.9 & \textbf{62.6} & \textbf{46.4} & 13.2 & \textbf{52.3} & 49.7 & \textbf{76.1} & \textbf{52.6}\\		
		\hline
		\multicolumn{2}{l}{\textbf{CiT}-multi-meta} \\
		\rowcolor{defaultcolor} ViT-B/16 & MoCo-v3 & 31 & 
        68.1 & 84.3 & 62.0 & 63.7 & 56.9 & 65.7 & 16.0 & 40.3 & 90.0 & 87.8 & 61.1 & 6.8 & 26.6 & 92.1 & 27.6 & 35.9 & \textbf{18.0} & 38.6 & 7.2 & 50.9 & 56.0 & 35.2 & 17.2 & 46.0 & 49.7 & 65.8 & 48.8\\
		ViT-B/32 & AugReg & 32 & 
        75.2 & 90.0 & 72.2 & 70.9 & 60.2 & 43.9 & 11.8 & 42.8 & 86.6 & 90.2 & 74.6 & 29.2 & 21.6 & 93.0 & 31.7 & 33.3 & 13.5 & \textbf{44.7} & 6.9 & 51.1 & 61.7 & 38.7 & 14.9 & 49.9 & 50.1 & 66.2 & 51.0\\
		ViT-B/16 & AugReg & 51 & 
        80.2 & 91.5 & 74.4 & \textbf{75.1} & 62.3 & 53.7 & 15.5 & 40.1 & 87.2 & 90.8 & 76.3 & 12.3 & 31.2 & 92.4 & 28.1 & 38.3 & 13.2 & 18.6 & 7.8 & 60.5 & 66.0 & 42.5 & 14.0 & 50.3 & 50.0 & 71.7 & 51.7\\
		ViT-L/16 & AugReg & 61 & 
        81.6 & \textbf{92.7} & \textbf{79.2} & 72.3 & 63.8 & 56.9 & 15.7 & 42.6 & 88.5 & 92.9 & 73.9 & \textbf{22.6} & 33.3 & 94.1 & 30.9 & 38.4 & 16.9 & 27.7 & 8.7 & \textbf{56.7} & 68.4 & 45.5 & \textbf{16.4} & 50.0 & 48.3 & 74.8 & 53.6 \\
		ViT-H/14 & SWAG & 54 & 
        \textbf{92.1} & 91.0 & 71.8 & 71.7 & \textbf{66.3} & \textbf{77.4} & \textbf{18.7} & \textbf{51.3} & \textbf{93.8} & \textbf{92.2} & \textbf{81.5} & 14.9 & \textbf{39.6} & \textbf{97.5} & \textbf{39.4} & \textbf{50.0} & 15.0 & 19.1 & \textbf{17.8} & 50.9 & \textbf{71.8} & \textbf{52.4} & 14.7 & \textbf{51.7} & \textbf{51.1} & \textbf{76.5} & \textbf{56.5}\\
	\end{tabular}
}
\caption{CiT trained on LAION400M and evaluated on 26 CLIP benchmarks: We vary metadata from IN-1K (CiT-1K-meta), IN-21K (CiT-21K-meta) and combined class names from 26 benchmarks (CiT-multi.-meta). 
	We also list results from CLIP on WIT400M and OpenCLIP trained on LAION400M. 
	*: from \url{https://github.com/LAION-AI/CLIP_benchmark}, with some results using VTAB benchmark evaluation/prompts.
}
\vspace{10pt}
\label{tbl:mt_laion}
\end{table*}

\subsubsection{Full Results on Raw Image-Text Crawl}

In Table \ref{tbl:mt_m2c2}, we show the result of CiT trained on our raw image-text crawl and evaluated on 26 benchmarks.
With a larger raw data source, we realize CiT takes more time to converge. We set b = 30000
for CiT-multi-meta and b = 60000 for CiT-21K-meta. The
trend is similar to LAION-400M but raw Image-Text Crawl is not cleaned for vision-language association. Similar as in Table \ref{tbl:mt_laion}, CiT-multi-meta is better than CiT-21K-meta, but the gap is larger. We expect better accuracy for longer training.

\begin{table*}[!htp]
\centering
\vspace{20pt}
\scalebox{0.66}{
	\setlength\tabcolsep{2.7pt}
 \begin{tabular}{l l c | c c c c c c c c c c c c c c c c c c c c c c c c c c | c}
		Vis. Encoder & Init. & Hrs 
		& \rotatebox[origin=l]{90}{Food-101} 
		& \rotatebox[origin=l]{90}{CIFAR10} 
		& \rotatebox[origin=l]{90}{CIFAR100} 
		& \rotatebox[origin=l]{90}{CUB}
		& \rotatebox[origin=l]{90}{SUN397}
		& \rotatebox[origin=l]{90}{Cars}
		& \rotatebox[origin=l]{90}{Aircraft}
		& \rotatebox[origin=l]{90}{DTD}
		& \rotatebox[origin=l]{90}{Pets}
		& \rotatebox[origin=l]{90}{Caltech-101}
		& \rotatebox[origin=l]{90}{Flowers}
		& \rotatebox[origin=l]{90}{MNIST}
		& \rotatebox[origin=l]{90}{FER-2013}
		& \rotatebox[origin=l]{90}{STL-10}
		& \rotatebox[origin=l]{90}{EuroSAT}
		& \rotatebox[origin=l]{90}{RESISC45}
		& \rotatebox[origin=l]{90}{GTSRB}
		& \rotatebox[origin=l]{90}{KITTI}
		& \rotatebox[origin=l]{90}{Country211}
		& \rotatebox[origin=l]{90}{PCAM}
		& \rotatebox[origin=l]{90}{UCF101}
		& \rotatebox[origin=l]{90}{Kinetics700}
		& \rotatebox[origin=l]{90}{CLEVR}
		& \rotatebox[origin=l]{90}{HatefulMemes}
		& \rotatebox[origin=l]{90}{SST2}
		& \rotatebox[origin=l]{90}{ImageNet}
		& \rotatebox[origin=l]{90}{Avg}\\
		\shline
		\hline
		\multicolumn{3}{l}{CLIP (WIT400M) \cite{radford2021learning}} \\
		ViT-B/32 & scratch & 458 & 
        84.4 & 91.3 & 65.1 & 37.8 & 63.2 & 59.4 & 21.2 & 44.5 & 87.0 & 87.9 & 66.7 & 51.9 & 47.3 & 97.2 & 49.4 & 60.3 & 32.2 & 39.4 & 17.8 & 58.4 & 64.5 & 47.8 & 24.8 & 57.6 & 59.6 & 63.2 & 56.9\\
		ViT-B/16 & scratch& 981 & 
        89.2 & 91.6 & 68.7 & 39.1 & 65.2 & 65.6 & 27.1 & 46.0 & 88.9 & 89.3 & 70.4 & 56.0 & 52.7 & 98.2 & 54.1 & 65.5 & 43.3 & 44.0 & 23.3 & 48.1 & 69.8 & 52.4 & 23.4 & 61.7 & 59.8 & 68.6 & 60.1\\
		ViT-L/14 & scratch & 6803 & 
        92.9 & 96.2 & 77.9 & 48.3 & 67.7 & 77.3 & 36.1 & 55.3 & 93.5 & 92.6 & 78.7 & 87.2 & 57.5 & 99.3 & 59.9 & 71.6 & 50.3 & 23.1 & 32.7 & 58.8 & 76.2 & 60.3 & 24.3 & 63.3 & 64.0 & 75.3 & 66.2\\
		\hline
		\shline
		\multicolumn{3}{l}{\textbf{CiT}-1K-meta} \\
		\rowcolor{defaultcolor} ViT-B/16 & MoCo-v3 & 39 & 
        29.0 & 86.0 & 56.5 & 17.6 & 41.3 & 12.4 & 5.8 & 25.7 & 83.8 & 77.0 & 10.6 & \textbf{10.8} & 24.9 & 95.1 & 22.3 & 20.8 & 6.8 & 35.6 & 4.2 & 50.8 & 27.7 & 20.5 & \textbf{17.2} & 48.9 & 50.1 & 68.4 & 36.5\\
		ViT-B/32 & AugReg & 69 & 
        42.8 & 92.2 & 70.5 & \textbf{22.1} & 49.0 & 11.4 & 5.5 & 27.0 & 83.8 & 81.1 & 16.5 & 8.2 & \textbf{32.5} & 94.3 & \textbf{29.4} & 22.2 & 8.5 & \textbf{39.1} & 4.9 & 51.3 & 37.6 & 26.7 & 16.4 & 48.0 & 50.1 & 67.8 & 40.0\\
            ViT-B/16 & AugReg & 72 & 
            43.9 & 92.1 & 73.4 & 20.4 & \textbf{50.0} & 10.9 & 4.5 & 31.3 & 84.6 & 83.0 & 18.8 & 7.1 & 21.5 & 96.2 & 23.3 & 22.4 & \textbf{11.2} & 29.4 & 5.2 & \textbf{52.3} & 41.9 & 29.4 & 17.0 & \textbf{50.6} & 50.1 & 74.9 & 40.2\\
		ViT-L/16 & AugReg & 105 & 
            47.8 & \textbf{95.4} & \textbf{76.0} & 18.5 & 49.4 & 11.4 & 5.6 & 30.9 & \textbf{84.7} & \textbf{83.7} & 22.4 & 6.4 & 25.6 & 96.8 & 24.7 & 29.7 & 8.9 & 36.3 & 5.3 & 50.9 & 45.9 & 31.0 & 16.3 & 46.5 & 50.1 & \textbf{77.5} & 41.4\\
		ViT-H/14 & SWAG & 43 & 
            \textbf{57.2} & 93.2 & 68.5 & 19.8 & 47.2 & \textbf{25.6} & \textbf{5.9} & \textbf{32.4} & 81.3 & 82.5 & \textbf{25.3} & 8.2 & 28.8 & \textbf{97.4} & 17.6 & \textbf{42.2} & 8.1 & 29.2 & \textbf{10.3} & 50.9 & \textbf{53.7} & \textbf{38.8} & 14.5 & 48.0 & \textbf{53.2} & 77.1 & \textbf{43.0}\\
		\hline
		\multicolumn{3}{l}{\textbf{CiT}-21K-meta} \\
		\rowcolor{defaultcolor} ViT-B/16 & MoCo-v3 & 134 & 
        57.1 & 87.1 & 60.3 & 57.1 & 54.0 & 10.5 & \textbf{6.0} & 37.0 & 84.6 & 82.8 & 59.9 & 9.8 & 26.8 & 96.8 & 31.8 & 30.8 & 8.3 & 41.2 & 7.4 & 59.9 & 37.9 & 25.9 & \textbf{20.8} & 48.2 & 50.1 & 62.8 & 44.4\\
		ViT-B/32 & AugReg & 148 & 
        64.4 & 93.2 & 71.7 & 49.5 & 56.8 & 10.8 & 5.7 & 35.4 & 76.2 & 85.8 & 60.9 & 9.5 & 29.1 & 95.4 & 27.1 & 25.2 & 9.3 & 39.8 & 7.7 & 51.3 & 45.8 & 32.1 & 14.1 & \textbf{51.3} & 50.1 & 62.2 & 44.6\\
		ViT-B/16 & AugReg & 161 & 
        70.0 & 93.6 & 75.9 & 58.2 & 59.9 & 11.7 & 5.2 & \textbf{37.7} & 74.9 & 89.3 & 61.7 & 9.8 & 32.6 & 97.9 & \textbf{29.5} & 29.4 & 11.2 & 40.9 & 9.0 & 51.1 & 49.6 & 36.1 & 13.6 & 48.9 & 50.1 & 69.4 & 46.8\\
		ViT-L/16 & AugReg & 228 & 
        71.7 & \textbf{96.0} & \textbf{78.7} & 56.7 & \textbf{62.4} & 12.2 & 5.9 & 37.4 & 77.0 & \textbf{90.6} & 65.3 & \textbf{14.6} & \textbf{37.6} & 98.3 & 27.8 & 34.0 & 8.5 & \textbf{34.0} & 9.4 & 44.2 & 54.7 & 39.0 & 15.5 & 47.9 & 50.1 & 72.6 & 47.8\\
		ViT-H/14 & SWAG & 310 & 
        \textbf{80.4} & 93.2 & 72.0 & \textbf{58.4} & 60.8 & \textbf{25.6} & 5.5 & 36.0 & \textbf{78.4} & 89.1 & \textbf{70.6} & 7.8 & 34.7 & \textbf{98.9} & 28.4 & \textbf{41.7} & \textbf{10.8} & 29.9 & \textbf{14.0} & \textbf{50.8} & \textbf{57.5} & \textbf{41.9} & 12.4 & 45.8 & \textbf{52.6} & \textbf{75.5} & \textbf{49.0}\\
		\hline
		\multicolumn{2}{l}{\textbf{CiT}-multi-meta} \\
		\rowcolor{defaultcolor} ViT-B/16 & MoCo-v3 & 91 & 
        70.4 & 88.8 & 61.1 & 60.1 & 59.0 & 63.2 & 24.5 & 38.4 & 90.2 & 85.5 & 66.5 & 9.8 & 32.0 & 96.6 & \textbf{35.4} & 39.0 & 9.5 & \textbf{35.8} & 10.2 & 50.3 & 48.7 & 33.4 & \textbf{17.1} & 43.8 & 50.1 & 66.1 & 49.4\\
		ViT-B/32 & AugReg & 62 & 
        72.7 & 92.9 & 71.0 & 51.0 & 58.9 & 30.9 & 10.9 & 36.3 & 86.6 & 87.4 & 67.5 & 9.8 & 36.3 & 94.5 & 29.1 & 29.4 & 8.5 & 33.4 & 8.6 & 54.9 & 51.6 & 36.3 & 14.8 & \textbf{49.2} & 50.0 & 64.4 & 47.6\\
		ViT-B/16 & AugReg & 62 & 
        81.3 & 94.0 & 76.6 & 65.2 & 62.2 & 44.1 & 17.9 & \textbf{41.3} & 90.0 & 90.6 & 74.9 & 9.8 & 35.3 & 97.5 & 34.6 & 36.5 & 13.1 & 34.4 & 10.4 & \textbf{56.8} & 57.6 & 41.3 & 13.4 & 50.6 & 50.1 & 71.9 & 52.0\\
		ViT-L/16 & AugReg & 62 & 
        82.4 & \textbf{96.1} & \textbf{79.2} & 62.4 & 64.1 & 44.5 & 15.8 & 41.2 & 89.3 & 91.3 & 74.9 & 9.8 & 34.7 & 98.2 & 27.9 & 38.7 & 8.9 & 33.4 & 11.1 & 55.9 & 61.3 & 44.0 & 11.9 & 48.9 & 50.1 & 74.4 & 51.9\\
		ViT-H/14 & SWAG & 203 & 
        \textbf{93.7} & 93.5 & 73.2 & \textbf{75.7} & \textbf{65.1} & \textbf{79.5} & \textbf{25.2} & 40.3 & \textbf{95.8} & \textbf{92.1} & \textbf{85.0} & \textbf{11.6} & \textbf{38.9} & \textbf{98.3} & 30.5 & \textbf{51.9} & \textbf{10.1} & 28.7 & \textbf{21.8} & 52.5 & \textbf{68.9} & \textbf{52.9} & 15.9 & 45.7 & \textbf{50.1} & \textbf{77.6} & \textbf{56.7}\\
	\end{tabular}
}
\caption{CiT trained on Raw Image-Text Crawl and evaluated on 26 CLIP benchmarks: We vary metadata from IN-1K (CiT-1K-meta), IN-21K (CiT-21K-meta) and combined class names from 26 benchmarks (CiT-multi.-meta). The budget $b=60000$ for IN-21K and $b=30000$ for combined class names. We also list results from CLIP on WIT400M. 
}
\vspace{10pt}
\label{tbl:mt_m2c2}
\end{table*}

\begin{table*}[t]
\centering
\scalebox{0.65}{
    \setlength\tabcolsep{2.7pt}
    \begin{tabular}{l | c c c c c c c c c c c c c c c c c c c c c c c c c c }
YFCC15M
& \rotatebox[origin=l]{90}{Food-101} 
& \rotatebox[origin=l]{90}{CIFAR10} 
& \rotatebox[origin=l]{90}{CIFAR100} 
& \rotatebox[origin=l]{90}{CUB}
& \rotatebox[origin=l]{90}{SUN397}
& \rotatebox[origin=l]{90}{Cars}
& \rotatebox[origin=l]{90}{Aircraft}
& \rotatebox[origin=l]{90}{DTD}
& \rotatebox[origin=l]{90}{Pets}
& \rotatebox[origin=l]{90}{Caltech-101}
& \rotatebox[origin=l]{90}{Flowers}
& \rotatebox[origin=l]{90}{MNIST}
& \rotatebox[origin=l]{90}{FER-2013}
& \rotatebox[origin=l]{90}{STL-10}
& \rotatebox[origin=l]{90}{EuroSAT}
& \rotatebox[origin=l]{90}{RESISC45}
& \rotatebox[origin=l]{90}{GTSRB}
& \rotatebox[origin=l]{90}{KITTI}
& \rotatebox[origin=l]{90}{Country211}
& \rotatebox[origin=l]{90}{PCAM}
& \rotatebox[origin=l]{90}{UCF101}
& \rotatebox[origin=l]{90}{Kinetics700}
& \rotatebox[origin=l]{90}{CLEVR}
& \rotatebox[origin=l]{90}{HatefulMemes}
& \rotatebox[origin=l]{90}{SST2}
& \rotatebox[origin=l]{90}{ImageNet}\\
\shline

\hline
\# of classes & 101 & 10 & 100 & 200 & 397 & 196 & 100 & 47 & 37 & 102 & 102 & 10 & 7 & 10 & 10 & 45 & 43 & 4 & 211 & 2 & 101 & 700 & 8 & 2 & 2 & 1000\\
\hline
\textsc{$t>0.55$}\\
\# pairs per class (k) & 5.32 & 16.64 & 9.27 & 3.28 & 5.63 & 0.81 & 4.35 & 3.12 & 3.66 & 6.71 & 6.51 & 11.9 & 5.43 & 18.21 & 8.59 & 10.57 & 2.22 & 23.63 & 2.33 & 0.53 & 4.28 & 3.9 & 20.96 & 5.48 & 0.66 & 3.69\\
total keep rates (\%) & 3.66 & 1.13 & 6.31 & 4.46 & 15.2 & 1.08 & 2.96 & 0.999 & 0.922 & 4.66 & 4.52 & 0.81 & 0.259 & 1.24 & 0.585 & 0.324 & 0.65 & 0.644 & 3.34 & 0.007 & 2.95 & 18.6 & 1.14 & 0.075 & 0.009 & 25.1\\
\hline
\end{tabular}
}
\caption{Statistics of YFCC15M (title and description) coverage on 26 tasks of CLIP evaluation: Low coverage could explain the root cause of the poor performance of zero-shot transfer (\eg Cars, PCAM, etc.).}
\label{tbl:taskfilter}
\end{table*}

\subsubsection{Additional Ablations}
\label{sec:extra_ablation}
This section extends ablations in Table~\ref{tbl:ablation} of the main paper to \textit{(i)} evaluation prompts and \textit{(ii)} training objectives.

\paragraph{Evaluation Prompts.}
We first verify the effects of LiT's extra prompts on CiT in Table \ref{tbl:ablation:tt}. We obtain a +0.2\% gain by adding them to the CLIP prompts.

\paragraph{Training Objective.}  We ablate the $\mathcal{L}_\text{img2txt}$ training objective which our approach uses (see \S3.2 of the main paper). In Table \ref{tbl:ablation:tt} we see that this variant provides a +0.2\% gain over CLIP's objective that also incorporates a text2img loss. 

\subsubsection{Early Detection of Task Coverage} 
One extra benefit of curation is being able to detect zero-shot transferability. Although existing scaled pre-trainings have huge success, the coverage of pre-training data distribution for downstream tasks is largely unknown. 
We discuss this coverage issue below.

\paragraph{Task Coverage}. We obtain the statistics of curated data (offline in Table \ref{tbl:ablation:stage}) for the 26 tasks and show it in Table \ref{tbl:taskfilter}.
We consider a sample with a maximum cosine similarity for one class as one sample belonging to that class/task. We note that this is a hard-matching which does not necessarily cover the full class to sample correlation. 
Breaking down YFCC15M for different tasks partially explains the low performance on some. For example, SST2 (a binary classification task) has low image-text pair matches, explaining the low performance (close to random) for all models.

\newpage
{\small
\bibliographystyle{ieee_fullname}
\bibliography{egbib}
}

\end{document}